\documentclass[review]{elsarticle}

\usepackage{lineno,hyperref}
\usepackage{algcompatible}
\usepackage{algorithm}
\usepackage[shortlabels]{enumitem}
\usepackage{epstopdf}
\usepackage{dsfont}
\usepackage{graphicx}
\usepackage{subfigure}
\usepackage[section]{placeins}
\usepackage{multirow}
\usepackage{hyperref}
\usepackage{amsthm}
\usepackage{capt-of}
\usepackage{amsfonts}
\usepackage{xcolor}
\usepackage{natbib} 
\usepackage{cool}
\usepackage{centernot}
\usepackage[T1]{fontenc}
\usepackage[left=2.5cm,right=2.5cm, bottom = 2.2cm, top = 2.2cm]{geometry}
\usepackage{xr-hyper}

\newcommand{\R}{{\mathbb{R}}}

\modulolinenumbers[5]

\journal{Journal of \LaTeX\ Templates}

\begin{document}

\begin{frontmatter}

\title{On the Relation between Prediction and Imputation Accuracy under Missing Covariates}

\author{Burim Ramosaj$^*$, Justus Tulowietzki, Markus Pauly}
\address{Faculty of Statistics \\
	 Institute of Mathematical Statistics and Applications in Industry\\
	 Technical University of Dortmund \\
	44227 Dortmund, Germany}


%\author[mysecondaryaddress]{\corref{mycorrespondingauthor}}
\cortext[mycorrespondingauthor]{\textit{Corresponding Author:} Burim Ramosaj \\
	\textit{Email address:} \texttt{burim.ramosaj@tu-dortmund.de} }
%\ead{burim.ramosaj@uni-ulm.de, markus.pauly@uni-ulm.de}

%\address[mymainaddress]{Helmholtzstrasse 20, 89079 Ulm, Germany}

\begin{abstract}
	Missing covariates in regression or classification problems can prohibit the direct use of advanced tools for further analysis. Recent research has realized an increasing trend towards the usage of modern Machine Learning algorithms for imputation. It originates from their capability of showing favourable prediction accuracy in different learning problems. In this work, we analyze through simulation the interaction between imputation accuracy and prediction accuracy in regression learning problems with missing covariates when Machine Learning based methods for both, imputation and prediction are used. In addition, we explore imputation performance when using statistical inference procedures in prediction settings, such as coverage rates of (valid) prediction intervals. Our analysis is based on empirical datasets provided by the UCI Machine Learning repository and an extensive simulation study. %Based on our (empirical) findings, imputation and prediction accuracy seem to be positively associated. %While Random Forest based prediction intervals were more or less robust towards larger missing rates, our simulation study indicated that the imputation of missing covariates resulted into larger prediction intervals  with more liberal coverage rates - more or less independent of the considered imputation method.\\
\end{abstract}


\end{frontmatter}

%\linenumbers

\section{Introduction}
The presence of missing values in data preparation and data analysis makes the usage of state-of-the art statistical methods difficult to apply. Seeking for a universal answer to such problems %to a potentially diverse audience 
was the main idea of \cite{rubin2004multiple}, who introduced multiple imputation. Through imputation, one provides data analysts (sequences) of completed datasets, based on which, various data analysis procedures can be conducted. An alternative to imputation is the usage of so-called data adjustment methods: statistical methods, that directly treat missing instances during training or parameter estimation such as the full-information-maximum-likelihood method (see e.g. \cite{enders2001performance}) or the expectation-maximization algorithm (c.f. \cite{horton1999maximum}). A big disadvantage of these methods is the expertise knowledge on theoretical model construction, where the likelihood function of parameters of interest, for example, needs to be adopted appropriately in order to account for missing information. Such examples can be found in \cite{amro2017permuting, amro2019multiplication, amro2021asymptotic}, e.g., where whole statistical testing procedures were adjusted to account for missing values. It is already well-known that more naive methods such as list-wise deletion or mean imputation can lead to severe estimation bias, see e.g. \cite{greenland1995critical, graham1996maximizing, jones1996indicator, chen2004nonparametric, rubin2004multiple}. Therefore, we do not discuss these approaches further. \\

In the current paper we focus on regression problems, where we do not have complete information on the set of covariates. 
%For most regression methods complete information on the set of included covariates is essential to conduct a wide range of methods. This, e.g. holds for ordinary least-square  maximum-likelihood estimation or the application of modern Machine Learning tools like the Random Forest or Boosting Machines. 
Missing covariates in supervised regression learning have been part in a variety of theoretical and applicative research fields. In \cite{chen2004nonparametric}, for example, a theoretical analysis based on maximum semiparametric likelihood for constructing consistent regression estimates was conducted. While in \cite{van1999multiple, yang2005imputation} or \cite{sterne2009multiple}, for example, multiple imputation is used as a tool in medical research for variable selection or bias-reduction in parameter estimation. More recent research has been focusing on Machine Learning (ML) based imputation. In \cite{stekhoven2012missforest, shah2014comparison, tang2017random, mayer2018package}, for example, the Random Forest method is used to impute missing values in various datasets with mixed variable scales. Other ML-based methods such as the $k$-nearest neighbor method, boosting machines or Bayesian Regression in combination with classification and regression trees have been part of (multiple) imputation, see e.g. \cite{chen2000nearest,xu2016sequential,dobler2017nonparametric,ramosaj2019predicting,zhang2019xgboost}.\\

Choosing an appropriate imputation method for missing data problems usually depends on several aspects such as the capability to treat mixed-type data, to impute reasonable values in variables of e.g. bounded support and to provide a fast imputation solution. Imputation accuracy can usually be assessed through the consideration of performance measures. Here, depending on the subsequent application, one may focus on data reproducibility measured through normalized root mean squared error ($NRMSE$) and proportion of false classification ($PFC$) or distribution preserving measures such as Kolmogorov-Smirnov based statistics, see e.g. \cite{stekhoven2012missforest, ramosaj2019predicting, thurow2021goodness}. 
 It is important to realize that these two classes of performance measures for the evaluation of imputation do not always agree, see, e.g., \cite{ramosaj2020cautionary, thurow2021goodness}. In fact, one could be provided by an imputation scheme with comparably low $NRMSE$ values, for which subsequent statistical inference can have highly-inflated type-I error rates. Therefore, choosing imputation methods by solely focusing on data reproducibility will not always lead to correct inference procedures. To control for the latter, \cite{rubin2004multiple} coined the term \textit{proper imputation}: a property, that guarantees correct statistical inference under the (multiple) imputation scheme. 
 
 While imputation accuracy can be accessed through data reproducibility or distributional recovery, prediction performance of subsequently applied ML methods (i.e. after imputation) are often evaluated using the mean-squared error ($MSE$) or the misclassification error ($MCE$). Under missing covariates, however, the sole focus on these measures is not sufficient, as the disagreement between data reproducibility and statistical inference has shown. In fact, beyond point-prediction we may also be interested in uncertainty quantifiation in form of prediction intervals. The effect of missing covariates on the latter remains mostly unknown. In this paper, we aim to close this gap in empirical and simulation-based scenarios. \\

We thereby put a special focus on the relation between data reproducibility and correct statistical inference for post-imputation prediction. While the latter could also be measured by distributional discrepancies like in \cite{thurow2021goodness}, we aim to put a special focus on coverage rates and prediction interval lengths in post-imputation prediction instead. The reason for this is the raise in ML-based imputation methods and their competitive predictive performance for both, imputation and prediction. Taking into account that data reproducibility and statistical inference are not always in harmony under the missing framework, an interesting question remains whether appropriate data reproducible imputation schemes lead to favourable prediction results after imputation. Furthermore, it is unknown whether imputation schemes with comparably low $NRMSE$ values will lead to accurate predictive machines in terms of delivering appropriate point predictions for future outcomes while also correctly quantifying their uncertainty. Therefore, based on different ML-methods in supervised regression learning problems, we (i) aim to clarify the interaction between $NRMSE$ and $MSE$ as measures for data reproducibility and predictive post-imputation ability. Furthermore, we (ii) aim to enlighten the impact of accurate data-reproducible imputation schemes on predictive, post-imputation statistical inference in terms of correct uncertainty quantification. To measure the latter, we take into account correct coverage rates and (narrow) interval lenghts of pointwise prediction intervals in post-imputation settings obtained through ML-methods.

\section{Measuring Accuracy}\label{Sec:Accuracy}

Measuring imputation accuracy can happen in many ways. A lot of research has been focused on the general idea of \textit{reconstructing} missing instances and being as \textit{close} as possible to the true underlying data (c.f. \cite{stekhoven2012missforest}, \cite{ramosaj2019predicting}). Although this approach seems reasonable, several disadvantages have been discovered when using data reproducible measures such as squared error loss, especially for later statistical inference; see e.g. \cite{rubin2004multiple,ramosaj2020cautionary,thurow2021goodness}. Therefore, we discuss several suitable measures for assessing prediction accuracy in regression learning problems with missing covariates. In the sequel, we assume that we have access to iid data collected in $\mathcal{D}_n  = \{ [\boldsymbol{X}_i^\top, Y_i]^\top \in \R^{p+1} : i = 1, \dots, n \}$, where 
\begin{align}\label{RegModel}
    Y_i = m(\boldsymbol{X}_i ) + \epsilon_i.
\end{align}
Here, $m(\boldsymbol{x}) = \mathbb{E}[Y_1 | \boldsymbol{X}_1 = \boldsymbol{x}]$ is the regression function, $\{\epsilon_i\}_{i = 1}^n$ is a sequence of iid random variables with $\mathbb{E}[\epsilon_1] = 0$ and $Var(\epsilon_1) = \sigma^2 \in (0, \infty)$
and we assume continuos covariates $X_i$. Missing positions in the features $\{ \boldsymbol{X}_i \}_{i = 1}^n$ are modeled by the indicator matrix $\boldsymbol{R} = \{ R_{ij} \}_{ij} \in \{0,1\}^{n \times p}$, where $R_{ij} = 0$ indicates that the $i$-th observation of feature $j \in  \{ 1,\dots,p \}$ is not observable. Focusing on the general issue of predicting outcomes in regression learning for new feature outcomes, we restrict our attention to data reproducible accuracy measures and model prediction accuracy measures. In order to cover statistical inference correctness for prediction, we use  ML-based prediction intervals as proposed in \cite{meinshausen2006quantile,zhang2019random,ramosaj2021interpretable} to account for coverage rates and interval lengths.

\subsection{Imputation and Prediction Accuracy}
In our setting, (missing) covariates are continuously scaled leading to the usage of accuracy measures for continuous random variables. Regarding imputation accuracy, we consider the $NRMSE$ formally given by 
\begin{align}
    NRMSE &= \frac{ \sqrt{ \sum\limits_{ (i, j) \in N_{mis} } ( X_{ij}^{imp} - X_{ij}^{mis} )^2 }}{ \sqrt{ \sum\limits_{(i, j) \in N_{mis} } ( X_{ij}^{imp} - \bar{X}_{\cdot \cdot}^{mis} )^2 } },
\end{align}
where $N_{mis} = \{ (i, j) \in \{1, \dots, n\} \times \{1, \dots, p\}  | R_{ij} = 0 \}$ is the set of all observations and features with missing entries in those positions. Here, $X_{ij}^{imp}$ denotes the imputed value of observation $i$ for variable $j$, while $X_{ij}^{mis}$ is the true, unobserved component of those positions. $X_{\cdot \cdot}^{mis}$ is the mean of the sequence $\{ X_{ij} : R_{ij} = 0 \}$.  Regarding the overall model performance on prediction, we make use of the mean squared error
\begin{align}
    MSE &= \mathbb{E}[ (Y - \widehat{m}_n(\boldsymbol{X}))^2 ], 
\end{align}
where $\widehat{m}_n$ is an ML-based estimator of $m$ on $\mathcal{D}_n$ and $[\boldsymbol{X}^\top, Y]^\top$ is an independent copy of $[\boldsymbol{X}_1^\top, Y_1 ]^\top$. Note that in the missing framework, $m$ is estimated on the imputed dataset $\mathcal{D}_n^{imp}$, while the $MSE$ is (usually) estimated using cross-validation procedures. 

\subsection{Prediction Intervals}
Based on the methods for uncertainty quantification proposed in \cite{meinshausen2006quantile,ramosaj2019consistent,zhang2019random,ramosaj2021interpretable}, we make use of Random Forest based prediction intervals. In an extensive simulation study in \cite{ramosaj2021interpretable}, it could be seen that other, ML-based prediction intervals such as the (stochastic) gradient tree boosting (c.f. \cite{friedman2002stochastic}) or the XGBoost method (c.f. \cite{chen2016xgboost}) did not perform well under completely observed covariates. Therefore, we restrict our attention to those already indicating accurate coverage rates in completely observed settings. Meinshausen's Quantile Regression Forest (see \cite{meinshausen2006quantile}), for example, delivers a pointwise prediction interval for an  unseen feature point $\boldsymbol{X} = \boldsymbol{x}$, which is formally given by 
\begin{align}\label{MeinshausenPI}
    PI_{QRF, n} &= [ \widehat{Q}_{n, \alpha/2}(\boldsymbol{x}); \quad  \widehat{Q}_{n, 1- \alpha/2}(\boldsymbol{x}) ], 
\end{align}
where $\widehat{Q}_{n, \alpha/2}(\boldsymbol{x}) = \inf\{ y | \widehat{F}_n(y | \boldsymbol{x}) \ge \alpha/2 \}$ and $\widehat{F}_n(y | \boldsymbol{x})$ is a Random Forest based estimator for the conditional distribution function $F(y|\boldsymbol{x})$ of $Y | \boldsymbol{X}  = \boldsymbol{x}$. Other prediction intervals based on the Random Forest are, e.g., given in \cite{zhang2019random} and \cite{ramosaj2021interpretable}. Following the same notation as in \cite{ramosaj2021interpretable}, we refer with $\widehat{m}_{n, M}(\boldsymbol{x})$ to a Random Forest prediction at $\boldsymbol{x}$, trained on $\mathcal{D}_n$ using $M$ decision trees, while $z_{1-\alpha}$ is the corresponding quantile of the standard normal distribution. We consider the same residual variance estimators as in \cite{ramosaj2021interpretable}, where $\widehat{\sigma}_{n, M}$ is the trivial residual variance estimate, $\widehat{\sigma}_{n, Mcorrect}$ is the residual variance estimate with finite-$M$ bias correction and $\widehat{\sigma}_{n, M; W}$ is a weighted residual variance estimator, see also \cite{ramosaj2019consistent}. Moreover, we denote with $\widehat{D}_{n, \alpha/2}^{RF}$ the empirical quantile of the Random Forest Out-of-Bag residuals. With this notation we obtain four more prediction intervals:
\begin{align}
    PI_{n, empQ}(\boldsymbol{x}) &= [ \widehat{m}_{n, M}(\boldsymbol{x}) + \widehat{D}_{n, \alpha/2}^{RF}; \quad  \widehat{m}_{n, M}(\boldsymbol{x}) + \widehat{D}_{n, 1 - \alpha/2}^{RF}], \label{RF_EmpQuant}\\
    PI_{n, ResVar}(\boldsymbol{x}) &= [ \widehat{m}_{n,M}(\boldsymbol{x}) - z_{ 1 - \alpha/2} \cdot \widehat{\sigma}_{n, M} ; \quad \widehat{m}_{n,M}(\boldsymbol{x}) + z_{ 1 - \alpha/2} \cdot \widehat{\sigma}_{n, M} ], \label{RF_ResVarSimple} \\
    PI_{n, Mcorrect }(\boldsymbol{x}) &= [ \widehat{m}_{n, M}(\boldsymbol{x}) - z_{1- \alpha/2} \cdot \widehat{\sigma}_{n, Mcorrect}; \quad  \widehat{m}_{n, M}(\boldsymbol{x}) + z_{1- \alpha/2} \cdot \widehat{\sigma}_{n, Mcorrect} ],\label{RF_MCorrect}\\
    PI_{n, weighted}(\boldsymbol{x}) &= [\widehat{m}_{n, M}(\boldsymbol{x}) - z_{1- \alpha/2} \cdot \widehat{\sigma}_{n, M; W}; \quad \widehat{m}_{n, M}(\boldsymbol{x}) + z_{1- \alpha/2} \cdot \widehat{\sigma}_{n, M; W}]. \label{RF_weighted}
\end{align}

For benchmarking we additionally consider a prediction interval that is build under the linear model assumption. Imputation accuracy in inferential prediction under missing covariates is then assessed by considering Monte-Carlo estimated coverage rates and interval lengths.% under a finite choice of Monte-Carlo iterations.

\section{Imputation and Prediction Models}\label{Sec:ImpModels}

We made use of the following state-of-the-art ML regression models for prediction
\begin{itemize}
    \item the Random Forest as implemented in the \textsf{R}-package \texttt{ranger} (c.f \cite{mayer2018package}),
    \item the (stochastic) gradient tree boosting (SGB) method from the \textsf{R}-package \texttt{gbm} (c.f. \cite{friedman2002stochastic}) and 
    \item the XGBoost method, also known as Queen of ML (c.f. \cite{Morde2019}), as implemented in the \textsf{R}-package \texttt{xgboost}. 
\end{itemize}
For each of them, we fit a prediction model to the (imputed) data. Both boosting methods rely on additive regression trees that are fitted sequentially using the principles of gradient descent for loss minimization. XGBoost, however, is slightly different by introducing extra randomization in tree construction, a proportional shrinkage on the leaf nodes and a clever penalization of trees. We refer to \cite{friedman2002stochastic,chen2016xgboost,friedman2017elements} for details on the concrete algorithms. For benchmarking, a linear model is trained as well.\\

Although several imputation models are available on various (statistical) software packages, we put a special focus on Random Forest based imputation schemes and the multivariate imputation using chained equations (MICE) procedure (c.f. \cite{van1999multiple}, \cite{stekhoven2012missforest}, \cite{ramosaj2019predicting}). The reasons for this are twofold, but both have roots in the same theoretical issue called \textit{congeniality}, see \cite{meng1994multiple} for a formal definition. First, using the same model for prediction and imputation yields to inferential congeniality; a  potential disagreement of both models, however, can result in uncongenial (multiple) imputation methods. The latter yields to invalid (multiple) imputation inference, as can be seen in \cite{fay1992inferences} or \cite{meng1994multiple}, for example. Secondly, focusing on Bayesian models for imputation such as the MICE procedure, is in line with the general framework of congeniality and the idea of (multiple) imputation. Although we do not directly compute point-estimates during the analysis phase, interesting quantities in our framework are Random Forest based prediction intervals and estimators of the $MSE$.\\

\texttt{missForest} in \textsf{R} is an iterative algorithm developed by \cite{stekhoven2012missforest}, that imputes continuous and discrete random variables using trained Random Forests on complete subsets of the data, and imputes missing values through prediction with the trained Random Forest model. The process iterates in imputing missing values until a pre-defined stopping criterion is met.  Similar to the \texttt{missForest} algorithm, we substituted the core learning method by other ML-based methods such as the SGB method (in the sequel referred to as the \texttt{gbm} for the algorithmic implementation) and the XGBoost method (in the sequel referred to with \texttt{xgboost} for the algorithmic implementation). Both methods are implemented in \textsf{R} using the same algorithmic framework as \texttt{missForest}, while substituting the Random Forest method with the SGB resp. XGBoost. That means that we train the SGB resp. XGBoost on (complete) subsets of the data and impute missing values through the prediction of the trained model in an iterative fashion.\\

MICE is a family of Bayesian imputation models developed in \cite{van2011mice,van2018flexible}. Under the normality assumption (i.e. MICE NORM), the method assumes a (Bayesian) linear regression model, where every parameter in that model is drawn from suitable priors. The predictive mean matching approach (MICE PMM) is similar to MICE NORM, but does not impute missing values through the prediction of those points using the Bayesian linear model, but randomly selects among observed points that are closest to the same model prediction as MICE NORM. In addition to these methods, MICE enables the implementation of Random Forest based methods, referred to as MICE RF, see, e.g., \cite{doove2014recursive}. The latter assumes a modified Random Forest, where additional randomization is applied compared to the \texttt{missForest}. For example, instead of simply predicting missing values through averaging observations in leaf nodes, the method randomly selects them. In addition, in the complete subset of the data determined for training the Random Forest, potential missing values are not initially imputed by mean or mode values, but by random draws among observed values. In the sequel, we refer to the algorithmic implementation in \textsf{R} of all these method using the terms \texttt{mice\textunderscore norm}, \texttt{mice\textunderscore pmm} and \texttt{mice\textunderscore rf}. \\

\section{Simulation Design}\label{Sec:SimDesign}

Our simulation design is separated in two parts. In the first part, empirical data from the UCI Machine Learning Repository covering regression learning problems are considered for the purpose of measuring imputation and prediction accuracy. The following five datasets are considered: 
\begin{enumerate}
    \item The \underline{\textbf{Airfoile Data}} consists of $(p+1) = 6$ variables measured in $n = 1,503$ observations, where the target variable is the scaled sound pressure level measured in decibel. The aim of this study conducted by NASA is to detect the impact of physical shapes of airfoils on the produced noise.
    \item In the \underline{\textbf{Concrete Data}}, $(p+1) = 7$ variables are measured in $n = 1,030$ observations. The target variable is the concrete compressive strength measured in MPa units.  
    \item The aim in the \underline{\textbf{QSAR Data}} is to predict aquatic toxicity for a certain fish species. It consists of $(p+1) = 9$ variables measured in $n = 546$ observations. 
    \item The \underline{\textbf{Real Estate Data}} has $(p+1) = 6$ variables and $n = 413$ observations. The aim is to build a prediction model for house price developments in the area of New Taipei City in Taiwan. 
    \item The \underline{\textbf{Power Plant Data}} consists of $n = 9,568$ observations with $(p+1) = 5$ variables. The actual dataset is much larger in terms of observations, but only the first $9,568$ are selected to speed the computations. The aim of this dataset is to predict the electric power generation of a water power-plant in Turkey. 
\end{enumerate}

For each dataset, missing values under the MCAR scheme have been inserted with $r \in \{0.1, 0.2, 0.3, 0.5 \}$ missing rates. Then, missing values have been (once) imputed with the imputation methods mentioned in Section \ref{Sec:ImpModels}. The whole process is iterated using $MC_{imp} = 500$ Monte-Carlo iterates. 
Based on each imputed dataset, all of the above mentioned prediction models are trained and their prediction accuracy is measured using a five-fold cross-validated $MSE$. Regarding hyper-parameter tuning of the various prediction models, we conducted a grid-search using a ten-fold cross-validation procedure with ten replications on the completely observed data, prior to the generation of missing values. This has been conducted using the \textsf{R}-function \texttt{trainControl} of the \texttt{caret}-package \cite{kuhn2015short}.\\

In the second part of our simulation study, synthetic data has been generated with missing covariates to detect the effect of imputation accuracy on prediction interval coverage rates. Here, we have focused on pointwise prediction intervals. For sample sizes $n \in \{100,500,1000\}$, regression learning problems of the form $\{  [ \boldsymbol{X}_i^\top, Y_i ]^\top \}_{i = 1}^n$ have been generated using a $p = 10$ dimensional covariate space and model $(\ref{RegModel})$, where $X_i \overset{\text{iid}}{\sim} \mathcal{N}_{p=10}(0,\Sigma)$ and $\epsilon_i \overset{\text{iid}}{\sim} \mathcal{N}(0,\sigma^2)$ are simulated independent of each other. Missing values have been inserted under the MCAR scheme using various missing rates $r_{PI} \in \{ 0.1, 0.2, 0.3\}$. Regarding the functional relationship between features and response, different regression functions with coefficient $\boldsymbol{\beta}_{0} = [2, 4, 2, -3, 1, 7, -4, 0 ,0, 0]^\top$ are used such as 
\begin{enumerate}
    \item a linear model: $m(\mathbf{x}_i) = \mathbf{x}_i^\top \boldsymbol{\beta}_0$,
    \item a polynomial model: $m(\mathbf{x}_i) = \sum\limits_{ j = 1}^p \beta_{0,j} x_{i,j}^j$, \label{PolyModel}
    \item a trigonometric model: $m(\mathbf{x}_i ) = 2\cdot\sin( \mathbf{x}_i^\top \boldsymbol{\beta}_0 + 2  )$, \label{SinModel}
    \item and a non-continuous model:
	    \begin{align*}
	        m(\mathbf{x}_i) &= 	\begin{cases}
            \beta_{0,1}x_{i,1} + \beta_{0,2}x_{i, 2} + \beta_{0,3} x_{i, 3}, &\text{ if } x_{i,3} > 0.5, \\
            \beta_{0,4}x_{i, 4} + \beta_{0,5}x_{i, 5} + 3 &\text{ if } x_{i,3} \le 0.5.
			\end{cases}
		\end{align*}\label{NonContModel}
\end{enumerate}
In order to capture potential dependencies among the features, various choices for the covariance matrix $\Sigma$ are considered: a positive auto-regressive, negative auto-regressive, compound symmetric, Toeplitz and the scaled identity structure. In addition, we aim to take care of the systematic variation originating from $m(\boldsymbol{X}_1)$, and the noise $\epsilon_1$, by choosing $\sigma^2$ in such a way that the signal-to-noise ratio $SN := Var(m(\boldsymbol{X}_1))/ \sigma^2  = 1$. Finally, using $MC_{PI} = 1,000$ Monte-Carlo iterations, prediction interval performance of the intervals proposed in Section \ref{Sec:Accuracy} are evaluated by approximating coverage rates and (average) interval lengths over the Monte-Carlo iterates.\\

\section{Simulation Results}\label{Sec:Results}
We present the results for each parts of the simulation study separately, starting with the empirical data sets with artificially inserted missing values. 
\subsection{Results on Imputation Accuracy and Model Prediction Accuracy}
In this section, we present the results for the empirical data analysis based on the Airfoile dataset using the imputation and prediction accuracy measures described in Section \ref{Sec:Accuracy} for evaluation. We thereby focus on the Random Forest and the XGBoost prediction model. The results of the linear and the SGB model as well as the results for all other datasets are given in the supplement (see Figures 1 - 19) and summarized at the end of this section.  %The detailed simulation results of the other prediction methods such as the linear model, the SGB model and the XGBoost model on the Airfoile dataset are shifted to the supplement (see Figures 1-3) and summarized at the end of this section. 

{\bf Random Forest as Prediction Model.} 
Figure~\ref{Airfoile_Plot_RF} summarizes for each imputation method the imputation error ($NRMSE$) and the model prediction error ($MSE$) over $MC_{imp} = 500$ Monte-Carlo iterates using the Random Forest method for prediction on the imputed dataset. On average, the smallest imputation error measured with the $NRMSE$ could be attained when using \texttt{missForest} and the \texttt{gbm} imputation method. In addition, these methods yielded low variations in $NRMSE$ across the Monte-Carlo iterates. In contrast, the  \texttt{mice\textunderscore norm}, \texttt{mice\textunderscore pmm} and \texttt{mice\textunderscore rf} behaved similarly resulting into largest $NRMSE$ values across the different imputation schemes with an increased variation in $NRMSE$ values. The \texttt{xgboost} method performed slightly worse than \texttt{missForest} and \texttt{gbm},  when focusing on imputation accuracy. In addition, all methods seem to be more or less robust towards an increased missing rate. Interesting is the fact that volatility decreases, as missing rates increase for the MICE procedures. Prediction accuracy measured in terms of cross-validated $MSE$ using the Random Forest model is lowest under the \texttt{missForest}, \texttt{xgboost} and \texttt{gbm}, which corresponds with the $NRMSE$ results. As expected, the estimated $MSE$ suffers from missing covariates and the effect gets worse with an increased missing rate. Hence, model prediction accuracy suffers from an increased amount of missing values, independent of the used imputation scheme. Although congeniality was defined for valid statistical inference procedures, the effect of using the same method for imputation and prediction seem to have also a positive effect on model prediction accuracy.\\

\begin{center}
	\centering
	\includegraphics[width=5in]{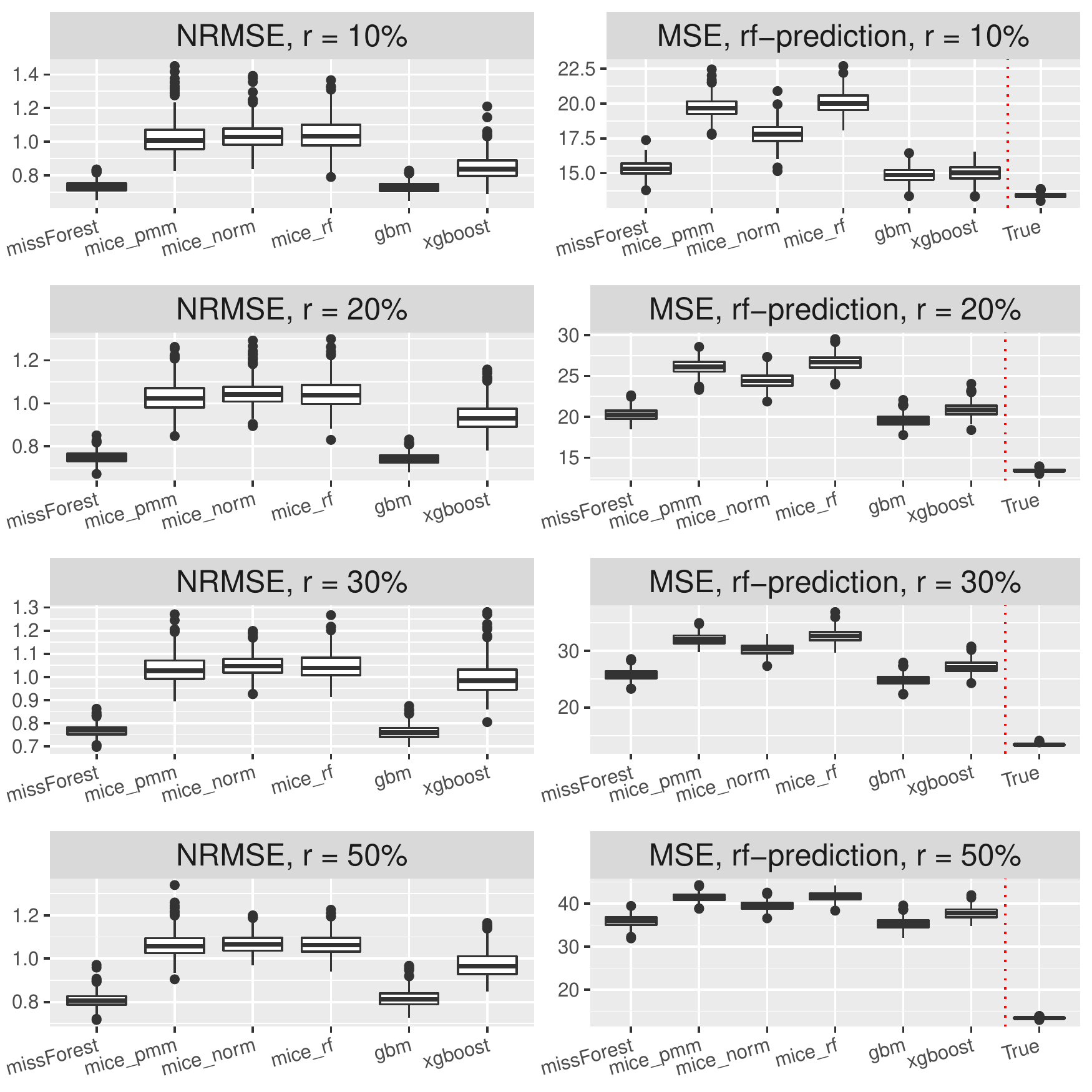}
	\captionof{figure}{ Imputation (left column) and prediction accuracy (right column) measured by $NRMSE$ and $MSE$ using the Random Forest method for predicting scaled sound pressure in the \textbf{Airfoile dataset} under various missing rates (listed row-wise). The $MSE$ is estimated based on a five-fold cross-validation procedure on the imputed dataset. The boxplots are over $500$ Monte-Carlo iterates. \textit{True} refers to the Airfoile dataset without any missing values and a fitted Random Forest on the complete dataset.}\label{Airfoile_Plot_RF}
\end{center}

{\bf XGBoost as Prediction Model.}
Switching the prediction method to XGBoost, one realizes an increase in model prediction accuracy for missing rates up to $20 \%$ as can be seen in Figure \ref{Airfoile_Plot_XGBoost}. In addition, for those missing rates, the \texttt{xgboost} imputation is competitive to the \texttt{missForest} method but looses in accuracy for larger missing rates compared to the \texttt{missForest}. Different to the Random Forest, the XGBoost prediction method is more sensitive towards an increased missing rate. Although under the completely observed framework the XGBoost method performed best in terms of estimated $MSE$, the results indicate that missing covariates can disturb the ranking. In fact, for missing rates $r \geq 30\%$, the Random Forest exhibits a better prediction accuracy. %A good indicator for this might be the $NRMSE$, that is computed resp. estimated prior to prediction model fitting.

 \begin{center}
	\centering
	\includegraphics[width=5in]{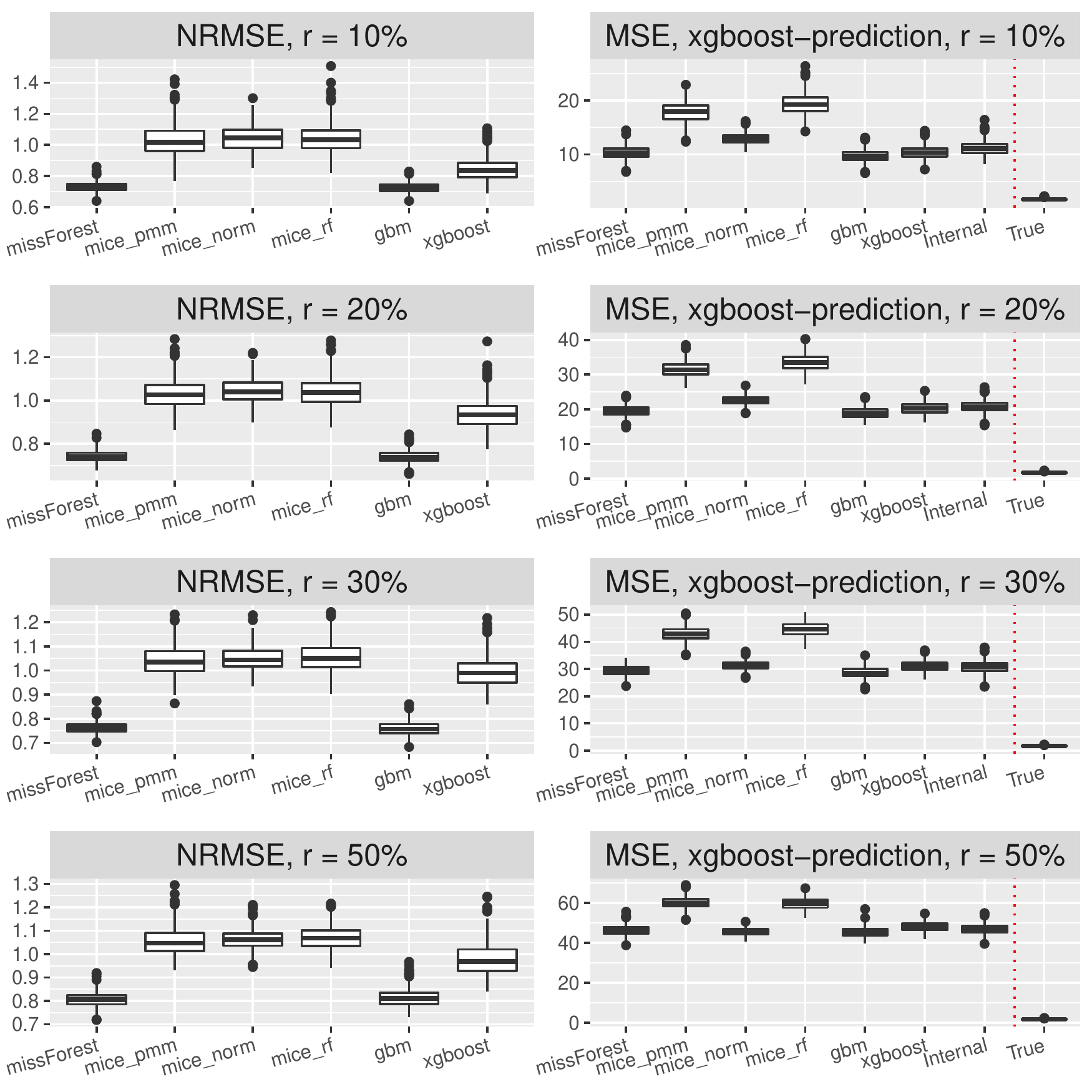}
	\captionof{figure}{ Imputation (left column) and prediction accuracy (right column) measured by $NRMSE$ and $MSE$ using the XGBoost method for predicting scaled sound pressure in the \textbf{Airfoile dataset} under various missing rates (listed row-wise). The $MSE$ is estimated based on a five-fold cross-validation procedure on the imputed dataset. The boxplots are over $500$ Monte-Carlo iterates. \textit{True} refers to the Airfoile dataset without any missing values and a fitted XGBoost on the complete dataset. \textit{Internal} is the internal missing value treatment of the XGBoost method without prior imputation. }\label{Airfoile_Plot_XGBoost}
\end{center}

{\bf Other Prediction models.} Using the linear model as the prediction model resulted in worse prediction accuracy with $MSE$ values ranging from $25$ ($r=10\%$) to $45$ ($r=50\%$). For all missing scenarios, using the \texttt{missForest} or the \texttt{gbm} method for imputation before prediction with the linear model 
resulted in the lowest $MSE$. %seems not to be as competitive as the Random Forest resp. XGBoost in the Airfoile and the other datasets (see Figures 5, 9, 13 and 17 \Justus{welche genau meinst du hier?} in the supplement). This can be seen in Figure 1 in the supplement. In total, similar results can be obtained as in Figure \ref{Airfoile_Plot_RF}, except that model prediction accuracy is worse. 
The results for the SGB method were even worse with $MSE$ values between $80$ and $99$. As a surprising  result prediction accuracy measured in terms of cross-validated $MSE$ decreased with increasing missing rate. A potential source of this effect could be the general weakness of the SGB in the Airfoile dataset without any missing values. After inserting and imputing missing values, which can yield to distributional changes of the data, it seems that the SGB method benefits from these effects. However, model prediction accuracy is still not satisfactory, see Figure 2 in the supplement.\\

%In addition, the MICE procedures such as \texttt{mice\textunderscore pmm} and \texttt{mice\textunderscore rf} where among the best method, which is completely different to the obtained results when using the Random Forest, linear model or the XGBoost method.  However, the SGB method yielded less accurate results in terms of $MSE$, even under fully observed covariates resulting into $MSE$ values between $82$ - $96$.  \\

{\bf Other Datasets.} For the other datasets, similar effects could be obtained. The Random Forest and the XGBoost showed the best prediction accuracy, see Figures 4 - 19 in the supplement. Again, larger missing rates affected model prediction accuracy for the XGBoost method, but the Random Forest was more robust to them overcoming XGBoost prediction performance measured in cross-validated $MSE$ for larger missing rates. Overall, $NRMSE$ and cross-validated $MSE$ seem to be positively associated to each other. Hence, more accurate imputation models seem to yield better model prediction measured by $MSE$.

\subsection{Results on Prediction Coverage and Length}

Using the prediction intervals in Section \ref{Sec:Accuracy}, we present coverage rates and interval lengths of pointwise prediction intervals in simulated data. Both quantities have been computed using $1,000$ Monte-Carlo iterations with sample sizes $n \in \{100,500,1000 \}$. The boxplots presented here (see Figures \ref{PI_Coverage_data_optimal_linear}, \ref{PI_Coverage_data_optimal_sinus}, \ref{PI_Length_data_optimal_linear} and \ref{PI_Length_data_optimal_sinus}) and in the supplement (see Figures 19 - 30) spread over the different covariance structures used during the simulation. Every row corresponds to one of the simulated missing rates $r \in \{ 0.1, 0.2, 0.3\}$, while the columns reflect the different Random Forest based prediction intervals. The left column summarizes the results for the Random Forest based prediction interval using empirical quantiles ($PI_{n, empQ}$), the center column reflects the Random Forest prediction interval using the simple residual variance estimator on Out-of-Bag errors ($PI_{n, ResVar}$), while the right column summarizes the Random Forest based prediction interval using the weighted residual variance estimator ($PI_{n, weighted}$). We shifted the results of $PI_{QRF, n}$, $PI_{n, MCorrect}$ and the prediction interval based on the linear model to the supplement (see Figures 19, 20, 21, 22, 24, 26, 28 and 30 in the supplement). Under the complete case scenarios, the latter methods did not show comparably well coverage rates as $PI_{n, emQ}$, $PI_{n, ResVar}$ and $PI_{n, weighted}$. For imputed missing covariates, the methods performed less accurate in terms of correct coverage rates, when comparing them with $PI_{n, emQ}$, $PI_{n, ResVar}$ and $PI_{n, weighted}$. Although the interval lengths of $PI_{QRF, n}$, $PI_{n, MCorrect}$ and the linear model were, on average smaller, coverage rate was not sufficient to make them competitive with $PI_{n, emQ}$, $PI_{n, ResVar}$ and $PI_{n, weighted}$. For prediction intervals that underestimated the $0.95$ threshold in the complete case scenario, we could observe more accurate coverage rates for larger missing rates. It seems that larger missing rates increase coverage rates for the $PI_{QRF,n}$ and $PI_{n, MCorrect}$ methods, independent of the used imputation scheme.

%In this section, we present the simulation results for all four models with $p = 10$ covariates and sample sizes of $n \in \{100,500,1000\}$. Figures \ref{PI_Coverage_data_optimal_linear} and \ref{PI_Coverage_data_optimal_sinus} show triple of row-wise boxplots which give the pointwise prediction interval coverage on the simulated data over all Monte-Carlo iterations based on the linear (Figure \ref{PI_Coverage_data_optimal_linear}) and trigonometric relationship (Figure Figure \ref{PI_Coverage_data_optimal_sinus} ). Each of the boxplots have the similar structure: The left column is based on the Random Forest PI using empirical quantiles, the center column on simple residual variance with Out-of-Bag errors and the right column using the weighted residual variance estimator. 

In Figure~\ref{PI_Coverage_data_optimal_linear} the boxplots of the linear regression model are presented. In general the usage of Random Forest based prediction intervals with empirical quantiles ($PI_{n, empQ}$) or simple variance estimation ($PI_{n, ResVar}$) show competitive behaviour in the complete case scenario. When considering the various imputation schemes, under the different missing rates, it can be seen that coverage rate slightly suffers compared to the complete case. To be more precise, larger missing rates lead to slightly larger coverage rates for $PI_{n, emQ}$, $PI_{n, ResVar}$ and $PI_{n, weighted}$. For the Random Forest based prediction interval with weighted residual variance, this effect seems to be positive, %since in general, the prediction intervals are more conservative leading to underestimated coverages. 
%Hence,
i.e. larger missing rates will lead to better coverage rates for $PI_{n, weighted}$. Comparing the results with the previous findings, we see that the \texttt{xgboost} yields, on average, the best coverage results across the different imputation schemes. While the MICE procedures did not reveal competitive performance in model prediction accuracy, the \texttt{mice\textunderscore norm} method under the linear model performed similar to the \texttt{missForest} procedure when comparing coverage rates. 

\begin{center}
	\centering
	\includegraphics[width=5in]{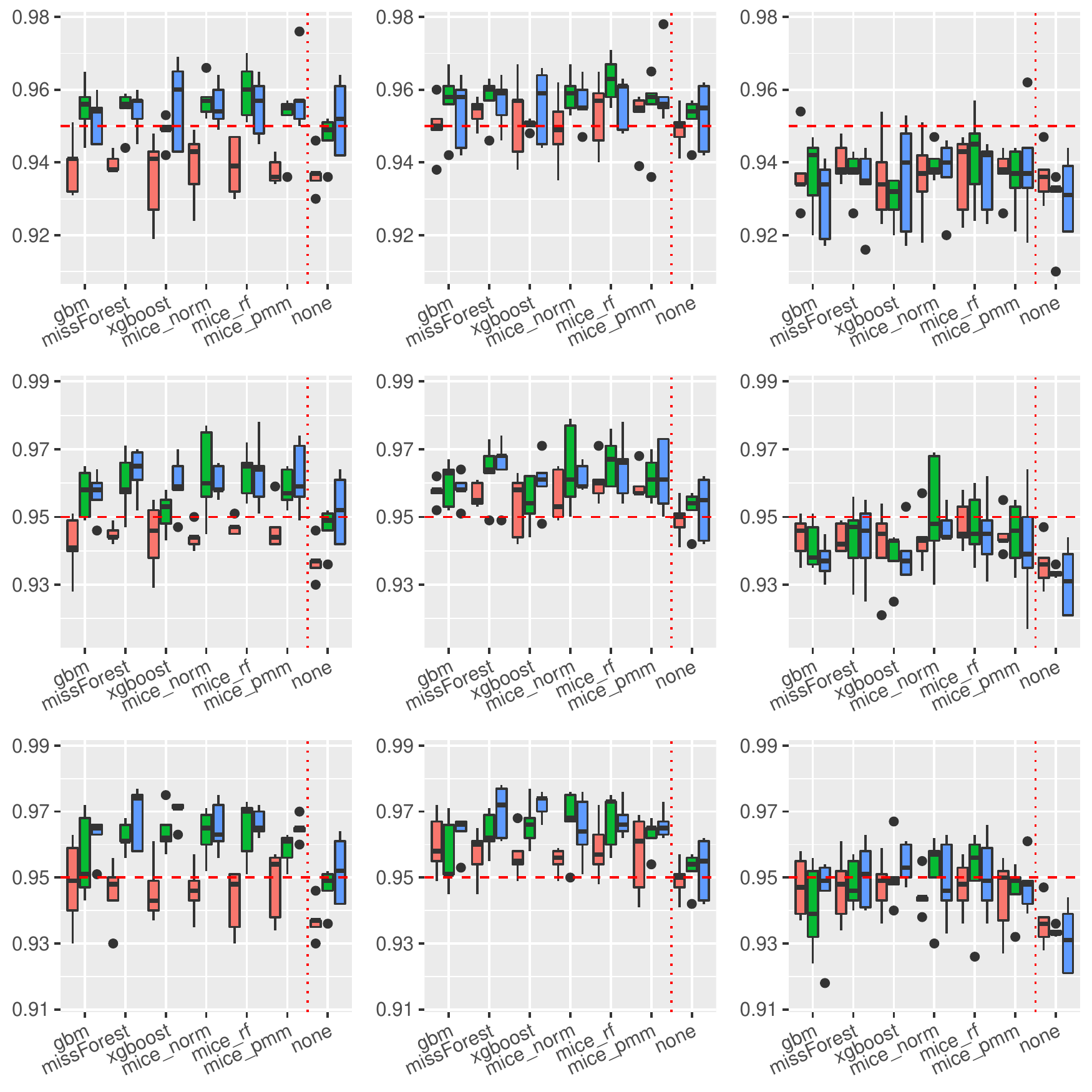}
	\captionof{figure}{ Boxplots of prediction  \textbf{coverage rates} under the \textbf{linear model}. The variation is over the different covariance structures of the features. Each row corresponds to one of the missing rates $r \in \{0.1, 0.2, 0.3\}$, while each column to the following prediction intervals: $PI_{n, empQ}$, $PI_{n, ResVar}$ and $PI_{n, weighted}$. The tripple (red, green and blue) correspond to the sample sizes $n \in (100,500,1000)$.} \label{PI_Coverage_data_optimal_linear}
\end{center}

Figure~\ref{PI_Coverage_data_optimal_sinus} summarizes coverage rates of pointwise prediction intervals under the trigonometric model. Similar to the linear case, all three methods $PI_{n, empQ}$, $PI_{n, ResVar}$ and $PI_{n, weighted}$ yield accurate coverage rates showing better approximation to the $0.95$ threshold when the sample size increase under the complete observation case. On average, the \texttt{xgboost} imputation method remains competitive compared to the other imputation methods. Slightly different to the linear case, the \texttt{mice\textunderscore norm} approach gains in correct coverage rate approximation compared to the \texttt{missForest}, together with the \texttt{mice\textunderscore pmm} approach. Nevertheless, the approximations between \texttt{mice\textunderscore norm},\texttt{mice\textunderscore pmm} and \texttt{missForest} are close to each other. As mentioned earlier, $PI_{n, weighted}$ turns more accurate in terms of correct coverage rates, when the missing rate increases. 

\begin{center}
	\centering
	\includegraphics[width=5in]{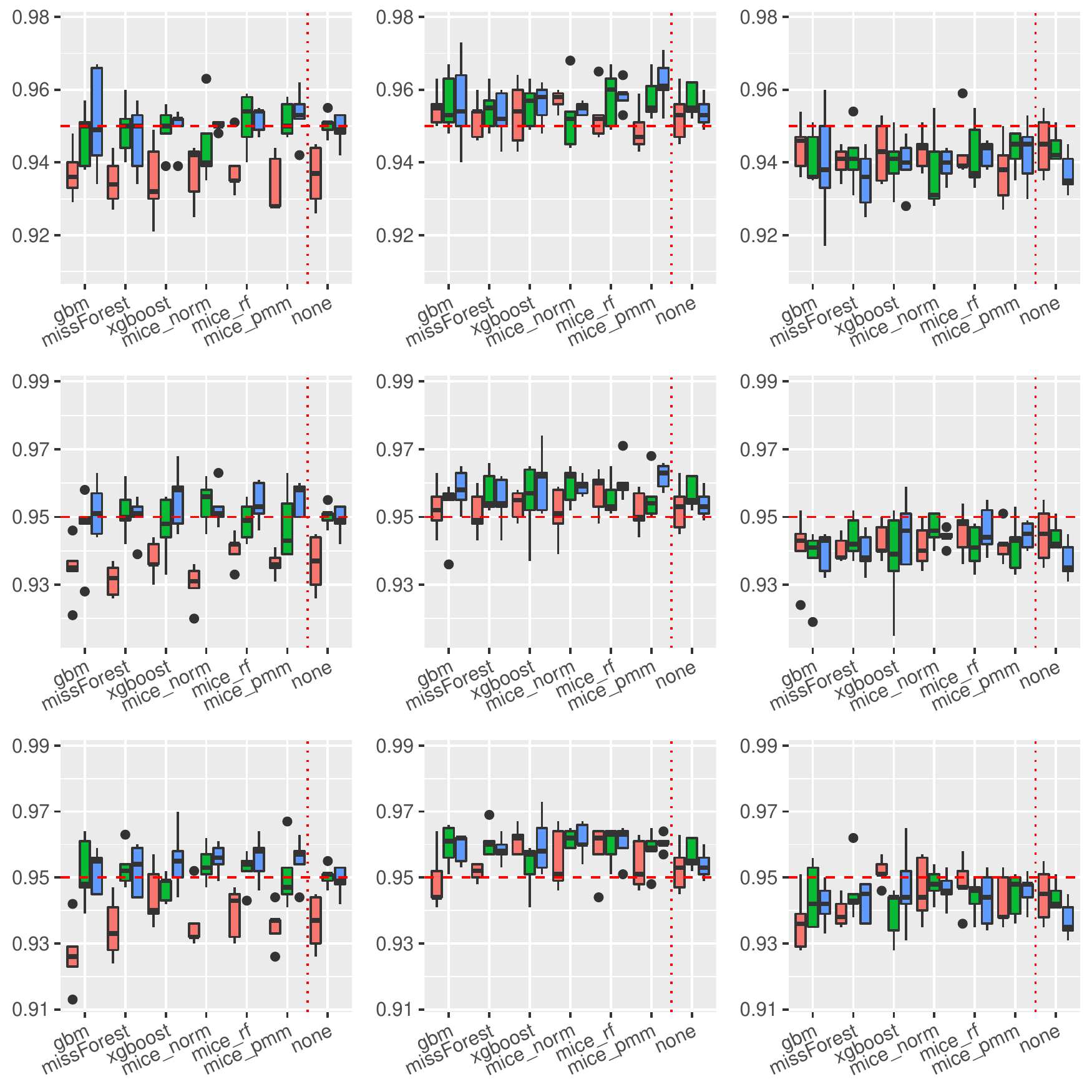}
	\captionof{figure}{ Boxplots of prediction  \textbf{coverage rates} under the \textbf{trigonometric model}. The variation is over the different covariance structures of the features. Each row corresponds to one of the missing rates $r \in \{0.1, 0.2, 0.3\}$, while each column to the following prediction intervals: $PI_{n, empQ}$, $PI_{n, ResVar}$ and $PI_{n, weighted}$. The tripple (red, green and blue) correspond to the sample sizes $n \in (100,500,1000)$.} \label{PI_Coverage_data_optimal_sinus}
\end{center}

Similar results compared to the linear and trigonometric case could be obtained for the polynomial model and the non-continuous model. Boxplots of the coverage rates can be found in Figures 24 and 28 of the supplement.  

Regarding the length of the intervals for the linear model (see Figure \ref{PI_Length_data_optimal_linear}), the prediction interval $PI_{n, empQ}$ and the parametric interval $PI_{n, ResVar}$ yield similar interval lengths. Under imputed missing covariates, however, the $PI_{n, empQ}$ interval leads to slightly smaller intervals than $PI_{n, ResVar}$. Nevertheless, the prediction interval based on the weighted residual variance estimator $PI_{n, wighted}$ had the smallest intervals on average. This comes with the cost of less accurate coverage rates as can be seen in Figure \ref{PI_Coverage_data_optimal_linear}. In addition, independent of the used prediction interval, an increased missing rate yields to larger intervals making the learning method such as the Random Forest more insecure about future predictions. Regarding the used imputation method, almost all imputation methods result into similar interval lengths. On average, the \texttt{missForest} method has slightly smaller intervals comparable to the \texttt{xgboost} imputation.

\begin{center}
	\centering
	\includegraphics[width=5in]{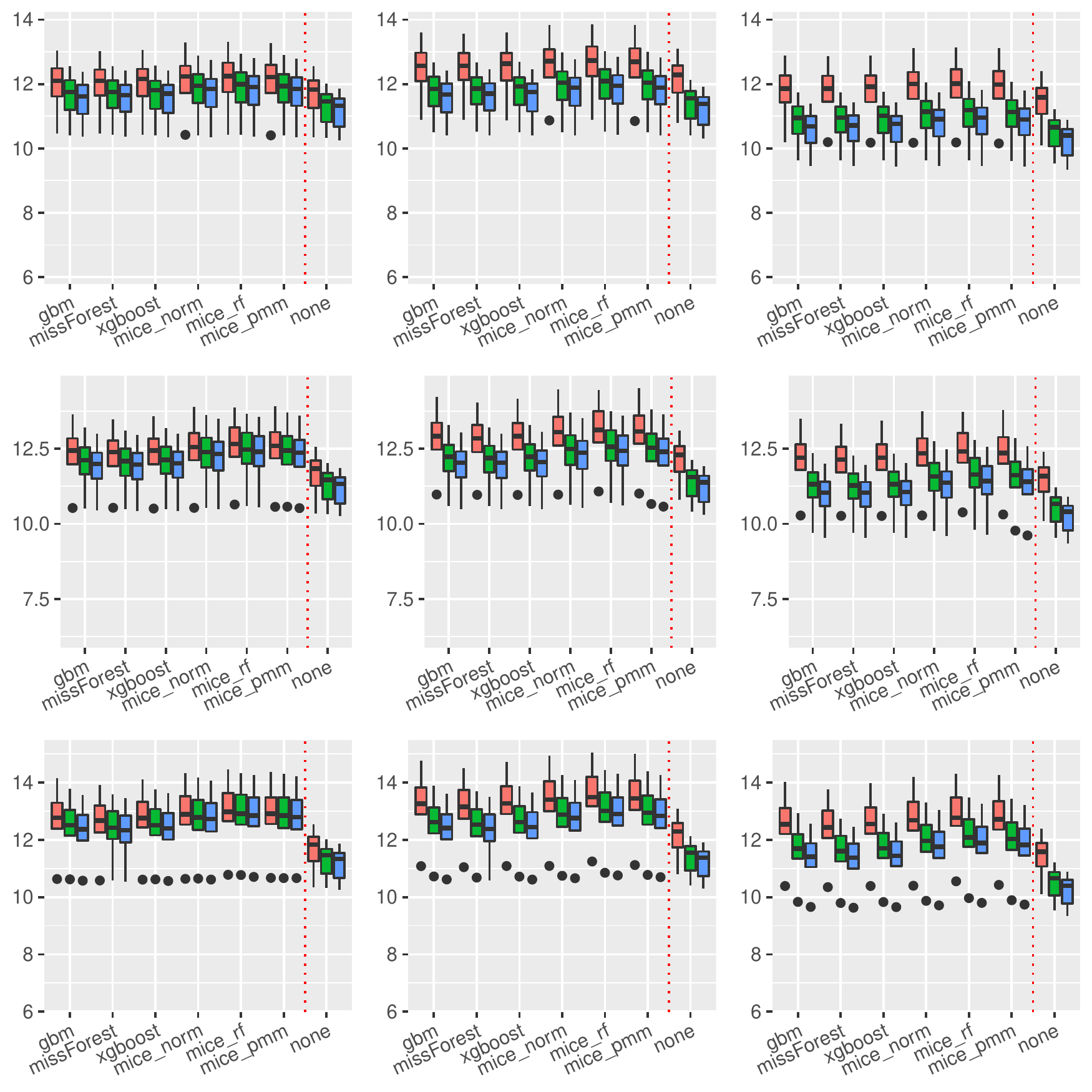}
	\captionof{figure}{Boxplots of prediction \textbf{interval lengths} under the \textbf{linear model}. The variation is over the different covariance structures of the features. Each row corresponds to one of the missing rates $r \in \{0.1, 0.2, 0.3\}$, while each column to the following prediction intervals: $PI_{n, empQ}$, $PI_{n, ResVar}$ and $PI_{n, weighted}$. The tripple (red, green and blue) correspond to the sample sizes $n \in (100,500,1000)$.}\label{PI_Length_data_optimal_linear}
\end{center}

Similar results on prediction lengths are obtained with other models. Considering the trigonometric function as in Figure \ref{PI_Length_data_optimal_sinus}, it can be seen that $PI_{n, empQ}$ results into slightly smaller intervals than $PI_{n, ResVar}$. However, the interval length for the empirical quantiles under the trigonometric model are more robust towards dependent covariates. 
Similar to the linear case, $PI_{n, weighted}$  results into the smallest interval lengths, but suffers from less accurate coverage. Furthermore, all imputation methods behave similar with respect to prediction interval lengths under the trigonometric case and other models (see Figures 21 and 22 in the supplement) . It can be seen that Random Forest based prediction intervals are, more or less, universally applicable to the different imputation schemes used in this scenario yielding similar interval lengths.   

\begin{center}
	\centering
	\includegraphics[width=5in]{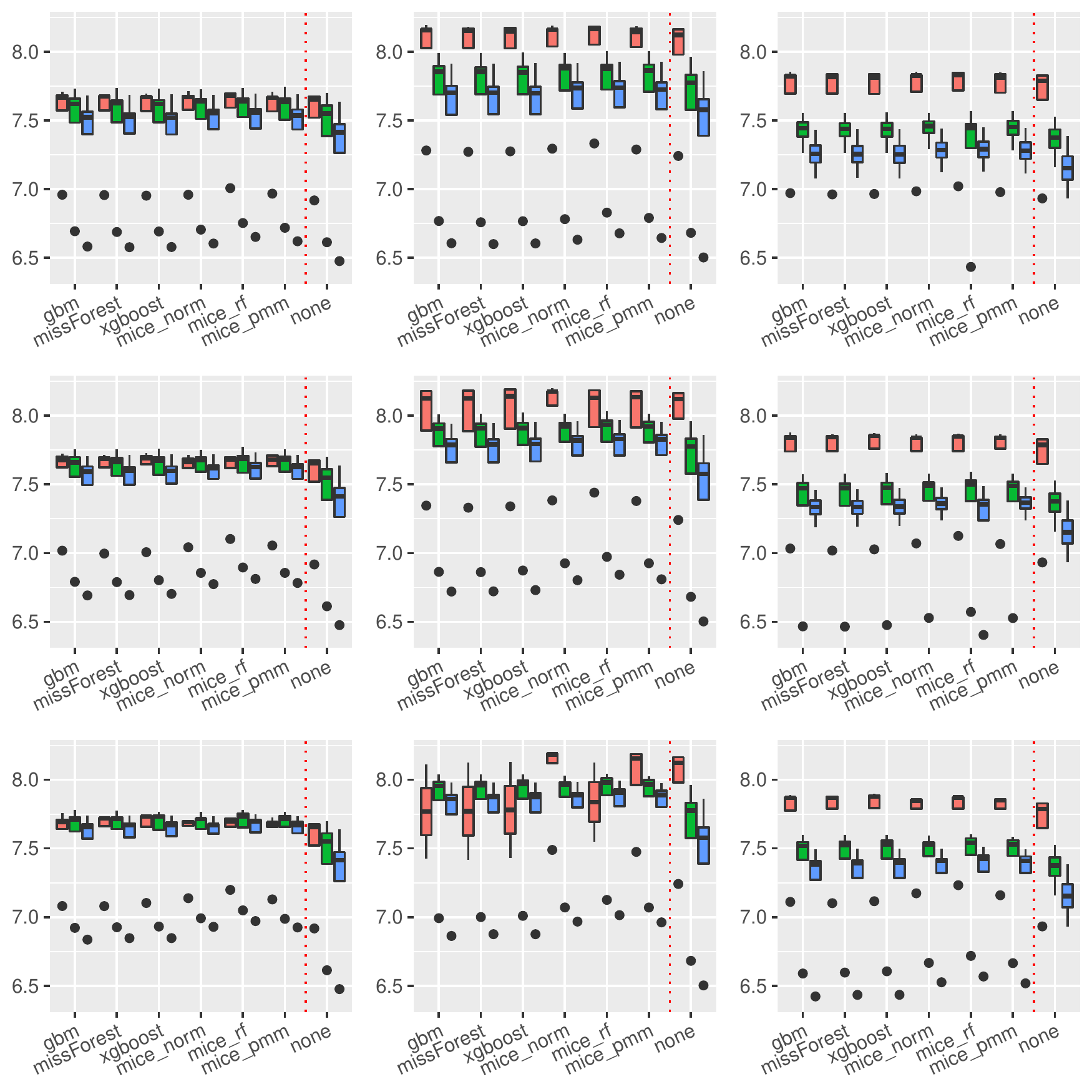}
	\captionof{figure}{Boxplots of prediction \textbf{interval lengths} under the \textbf{trigonometric model}. The variation is over the different covariance structures of the features. Each row corresponds to one of the missing rates $r \in \{0.1, 0.2, 0.3\}$, while each column to the following prediction intervals: $PI_{n, empQ}$, $PI_{n, ResVar}$ and $PI_{n, weighted}$. The tripple (red, green and blue) correspond to the sample sizes $n \in (100,500,1000)$.}\label{PI_Length_data_optimal_sinus}
\end{center}

In summary, Random Forest based prediction intervals with imputed missing covariates yield slightly wider intervals compared to the regression framework without missing values. For prediction intervals that are underestimating the true coverage rate, such as $PI_{QRF, n}$, $PI_{n, MCorrect}$ and $PI_{n, weighted}$, an increased missing rate had positive effects on the coverage rate. Overall, \texttt{missForest} and \texttt{xgboost} were competitive imputation schemes when considering accurate coverage rates and interval lengths using the $PI_{n, empQ}$ and $PI_{n, ResVar}$ intervals. \texttt{mice\textunderscore norm} resulted into similar, but slightly less accurate coverages compared to \texttt{missForest} and \texttt{xgboost}

\section{Conclusions}

Missing covariates in regression learning problems are common issues in data analysis problems. Providing a general approach for enabling the application of various analysis models is often  obtained through imputation. The use of ML-based methods in this framework has obtained an increased attention during the last decade, since fast and easy to use ML methods such as the Random Forest may provide us with quick and accurate fixes of data analysis problems. In our work, we put a special focus on variants of ML-based methods for imputation, that mainly rely on decision trees as base learners and their aggregation is conducted in a Random Forest based fashion or a boosting approach. We aimed to shed light into the general issue when and which imputation method should be used for missing covariates in regression learning problems that provide accurate point predictions with correct uncertainty ranges. To provide an answer to this, we conducted empirical analyses and simulations which are led by the following questions: 
Does an imputation scheme with a low imputation error (measured with the $NRMSE$) automatically provide us with accurate model prediction performance (in terms of cross-validated $MSE$)? 
How do ML-based imputation methods perform in estimating uncertainty ranges for future prediction points in form of pointwise prediction intervals? 
Are the results in harmony; that is, does an accurate imputation method with a low $NRMSE$ provide us with good model prediction accuracy measured in $MSE$ while delivering accurate and narrow prediction interval lengths?\\

By analyzing empirical data from the UCI Machine Learning repository, we found out that imputation methods with low imputation error measured with the $NRMSE$ yield better model prediction measured by cross-validated $MSE$. Especially for larger missing rates, the usage of the same ML method for both, imputation and prediction was beneficial. This is in line with the \textit{congeniality} assumption in \cite{meng1994multiple}; a theoretical term that (partly) guarantees correct inference after (multiple) imputation. In particular the \texttt{missForest} and our modified \texttt{xgboost} method for imputation yielded preferable results in terms of low imputation error and good model prediction. It is expected that ML methods with accurate model prediction capabilities measured in $MSE$ can be transformed to be used as an imputation method yielding low imputation errors, too. Regarding statistical inference procedures in prediction settings such as the construction of valid prediction intervals, Random Forest based imputation schemes such as the \texttt{missForest} and the \texttt{xgboost} method yielded competitive coverage rates and interval lengths. In addition, the MICE procedure with a Bayesian linear regression and normal assumption was under the aspect of correct coverage rates and interval lengths, competitive too. However, the method did not reveal low imputation error and overall good model prediction. Hence, based on our findings, the \texttt{missForest} and the \texttt{xgboost} method in combination with Random Forest based prediction intervals using empirical quantiles resp. Out-of-Bag estimated residual variances are competitive in three aspects: providing low imputation errors measured with the $NRMSE$, yielding comparably low model prediction errors measured by cross-validated $MSE$ and providing comparably accurate prediction interval coverage rates and narrow widths using Random Forest based intervals. Regarding the latter, our results also indicate that these intervals are competitively applicable to a wide range of imputation schemes.\\
Future research will focus on a theoretical exploration of the interaction between the $NRMSE$ and $MSE$ and the effect of the considered imputation methods on uncertainty estimators. Insights into their theoretical interaction will provide additional information to the general issue that \textit{imputation is not just prediction.}

%\section*{References}

\section*{Acknowledgment}
Burim Ramosaj's work is funded by the Ministry of Culture and Science of the state of NRW (MKW NRW) through the research grand programme KI-Starter. The authors gratefully acknowledge the computing time provided on the Linux HPC cluster at Technical University Dortmund (LiDO3), partially funded in the course
of the Large-Scale Equipment Initiative by the German Research Foundation (DFG) as project 271512359.

\bibliography{BibFile}

\end{document}

\begin{frontmatter}

\title{Supplementary Material to: On the Relation between Prediction and Imputation Accuracy under Missing Covariates}

\author{Burim Ramosaj$^*$, Justus Tulowietzki, Markus Pauly}
\address{Faculty of Statistics \\
	 Institute of Mathematical Statistics and Applications in Industry\\
	 Technical University of Dortmund \\
	44227 Dortmund, Germany} 
	


%\author[mysecondaryaddress]{\corref{mycorrespondingauthor}}
\cortext[mycorrespondingauthor]{\textit{Corresponding Author:} Burim Ramosaj \\
	\textit{Email address:} \texttt{burim.ramosaj@tu-dortmund.de} }
%\ead{burim.ramosaj@uni-ulm.de, markus.pauly@uni-ulm.de}

\end{frontmatter}

	\noindent This supplementary material includes additional simulation results belonging to the article \textit{On the Relation between Prediction and Imputation Accuracy under Missing Covariates}. Especially the results on imputation accuracy and model prediction accuracy based on the Airfoile, Concrete, QSAR, Real Estate and Power Plant dataset are presented. In addition, the results on prediction interval coverage and length regarding the four different models: linear, trigonometric, polynomial and non-continuous model are presented, too. 
	
	%Add here the table of content 
	\tableofcontents
	\newpage
	
\section{Results on Imputation Accuracy and Model Prediction Accuracy}
\subsection{Airfoile dataset}

\begin{center}
	\centering
	\includegraphics[width=5in]{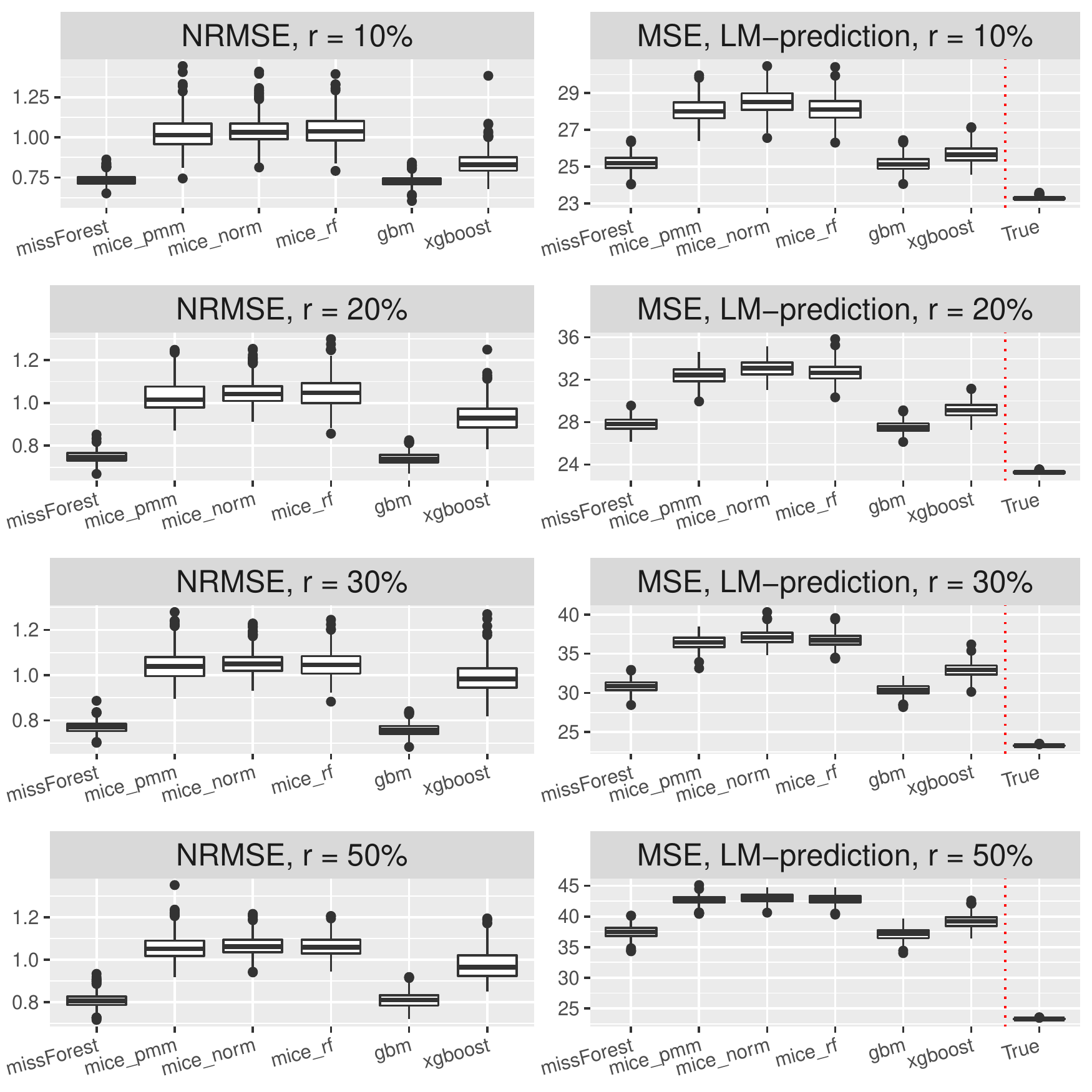}
	\captionof{figure}{ Imputation (left column) and prediction accuracy (right column) measured by $NRMSE$ and $MSE$ using a linear model for predicting scaled sound pressure in the Airfoile dataset under various missing rates (listed row-wise). The $MSE$ is estimated based on a five-fold cross-validation procedure on the imputed dataset. The boxplots are over $500$ Monte-Carlo iterates. \textit{True} refers to the Airfoile dataset without any missing values and a fitted Random Forest on the complete dataset.}\label{Airfoile_Plot_linear}
\end{center}

\begin{center}
	\centering
	\includegraphics[width=5in]{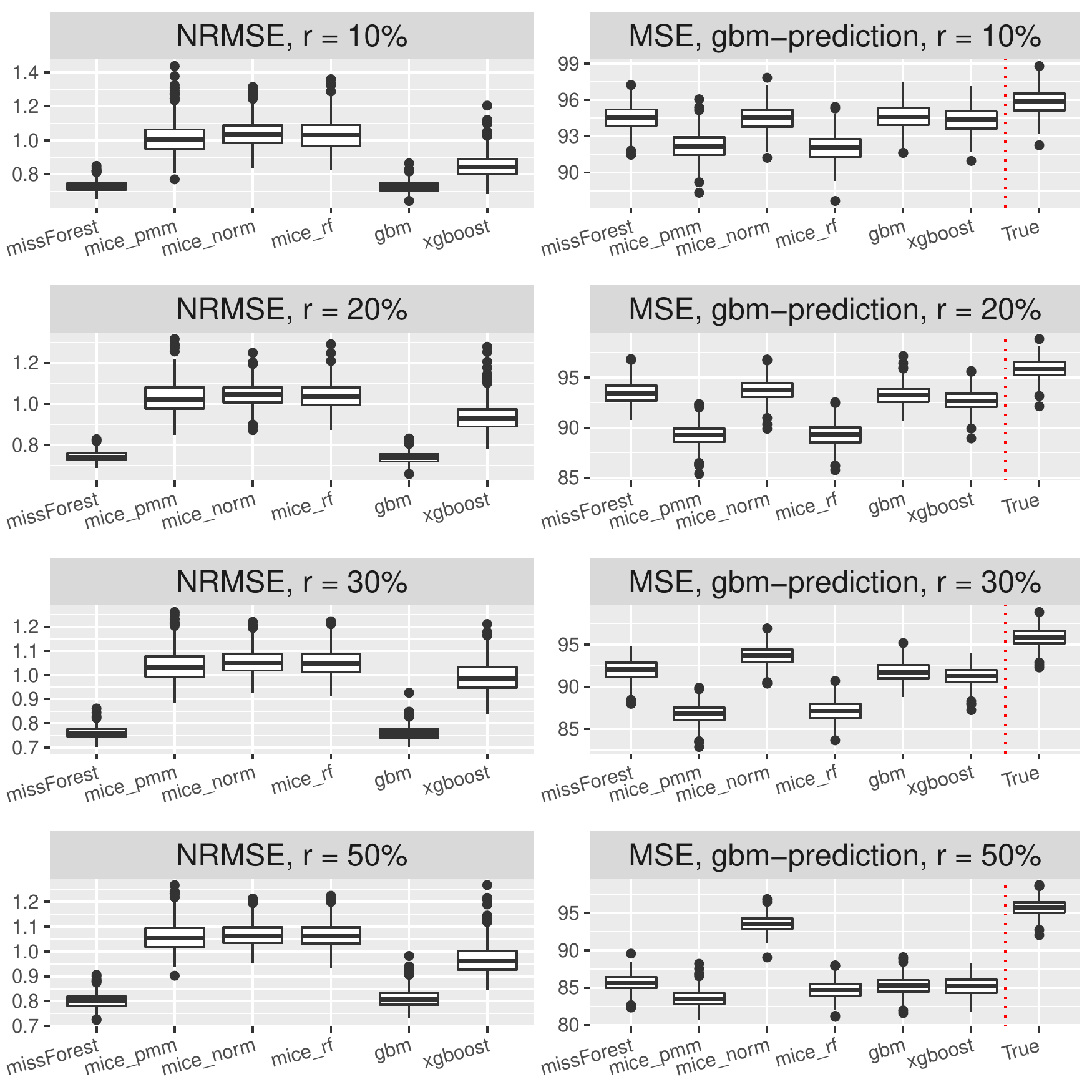}
	\captionof{figure}{  Imputation (left column) and prediction accuracy (right column) measured by $NRMSE$ and $MSE$ using the SGB method for predicting scaled sound pressure in the Airfoile dataset under various missing rates (listed row-wise). The $MSE$ is estimated based on a five-fold cross-validation procedure on the imputed dataset. The boxplots are over $500$ Monte-Carlo iterates. \textit{True} refers to the Airfoile dataset without any missing values and a fitted Random Forest on the complete dataset.}\label{Airfoile_Plot_gbm}
\end{center}

%\begin{center}
%	\centering
%	\includegraphics[width=5in]{./Pictures/Airfoile_Plot_Pred_xgboost.eps}
%	\captionof{figure}{Imputation (left column) and prediction accuracy (right column) measured by $NRMSE$ and $MSE$ using the XGBoost method for predicting scaled sound pressure in the Airfoile dataset under various missing rates (listed row-wise). The $MSE$ is estimated based on a five-fold cross-validation procedure on the imputed dataset. The boxplots are over $500$ Monte-Carlo iterates. \textit{True} refers to the Airfoile dataset without any missing values and a fitted Random Forest on the complete dataset.}\label{Airfoile_Plot_xgboost}
%\end{center}

\subsection{Concrete dataset}

\begin{center}
	\centering
	\includegraphics[width=5in]{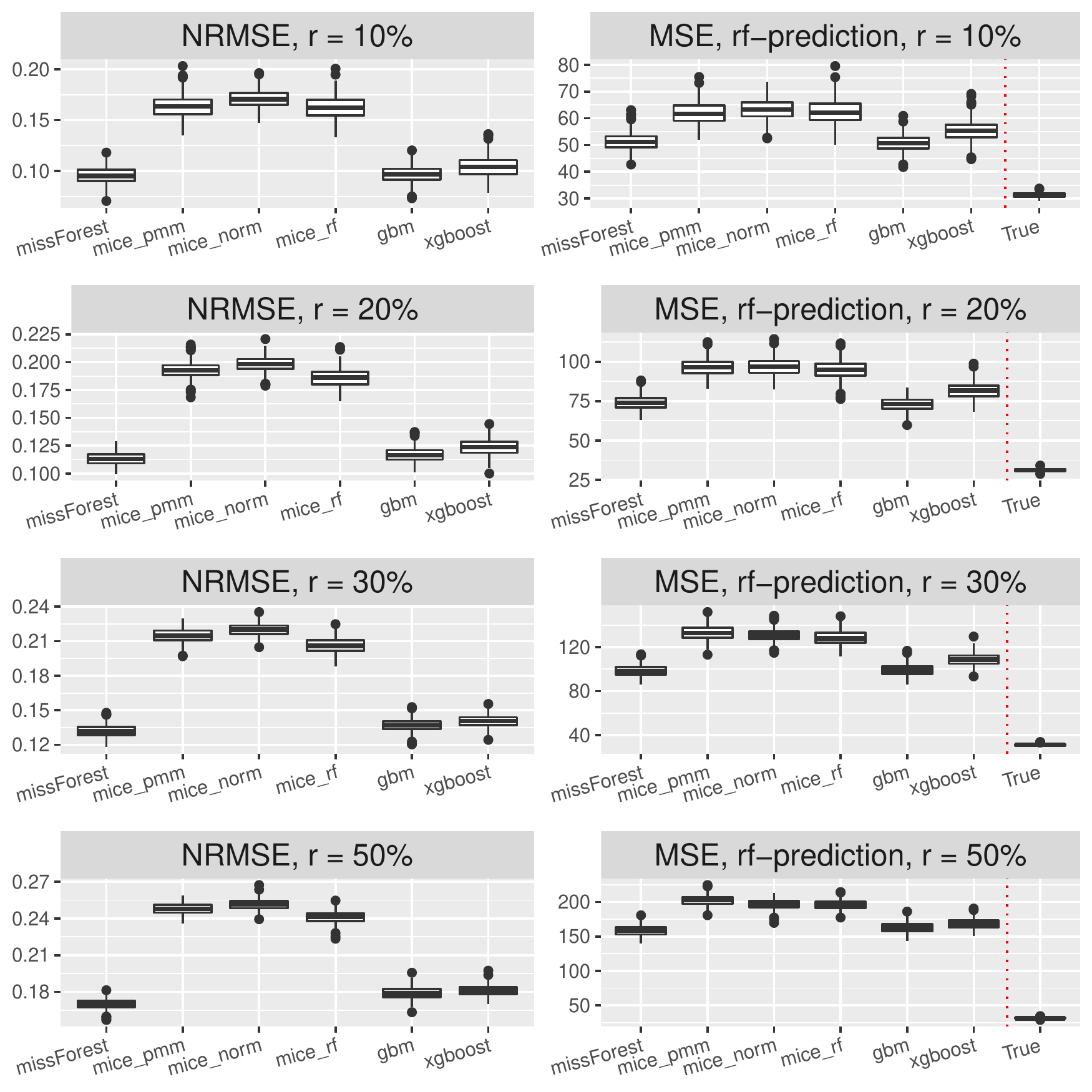}
	\captionof{figure}{Imputation (left column) and prediction accuracy (right column) measured by $NRMSE$ and $MSE$ using the Random Forest method for predicting scaled sound pressure in the Concrete dataset under various missing rates (listed row-wise). The $MSE$ is estimated based on a five-fold cross-validation procedure on the imputed dataset. The boxplots are over $500$ Monte-Carlo iterates. \textit{True} refers to the Concrete dataset without any missing values and a fitted Random Forest on the complete dataset. }\label{Concrete_Plot_RF}
\end{center}

\begin{center}
	\centering
	\includegraphics[width=5in]{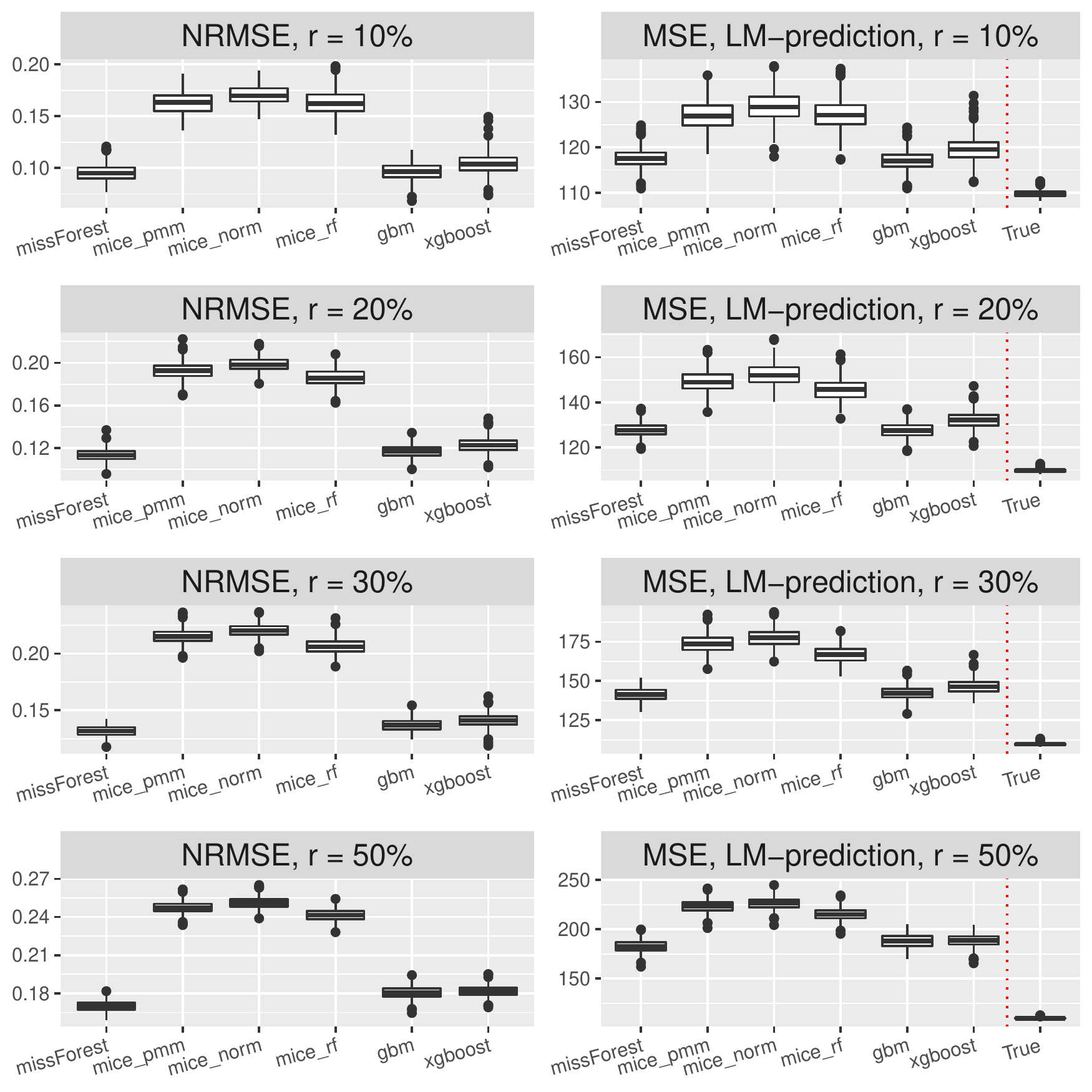}
	\captionof{figure}{Imputation (left column) and prediction accuracy (right column) measured by $NRMSE$ and $MSE$ using a linear model for predicting scaled sound pressure in the Concrete dataset under various missing rates (listed row-wise). The $MSE$ is estimated based on a five-fold cross-validation procedure on the imputed dataset. The boxplots are over $500$ Monte-Carlo iterates. \textit{True} refers to the Concrete dataset without any missing values and a fitted Random Forest on the complete dataset.}\label{Concrete_Plot_Pred_linear}
\end{center}

\begin{center}
	\centering
	\includegraphics[width=5in]{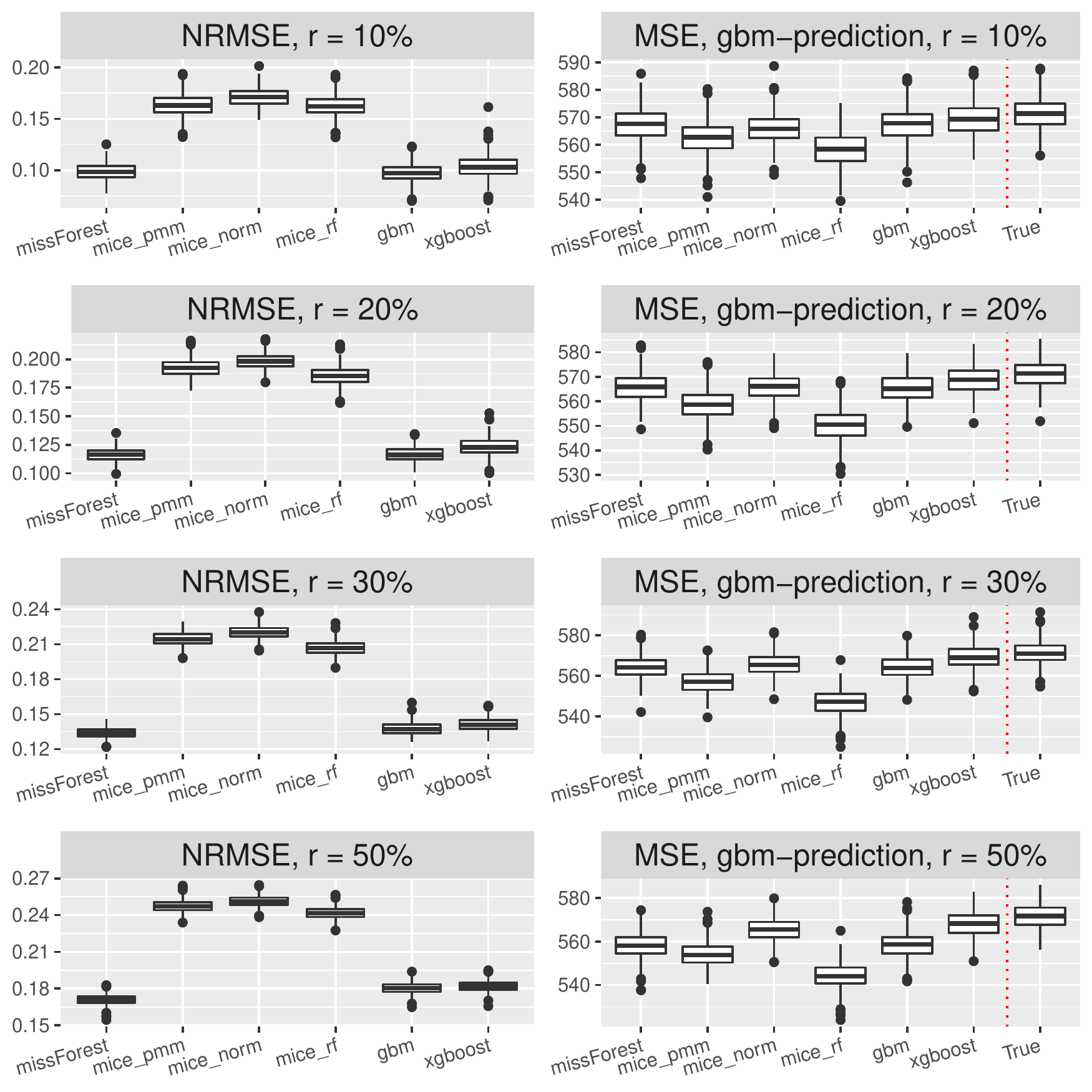}
	\captionof{figure}{Imputation (left column) and prediction accuracy (right column) measured by $NRMSE$ and $MSE$ using the SGB method for predicting scaled sound pressure in the Concrete dataset under various missing rates (listed row-wise). The $MSE$ is estimated based on a five-fold cross-validation procedure on the imputed dataset. The boxplots are over $500$ Monte-Carlo iterates. \textit{True} refers to the Concrete dataset without any missing values and a fitted Random Forest on the complete dataset.}\label{Concrete_Plot_gbm}
\end{center}

\begin{center}
	\centering
	\includegraphics[width=5in]{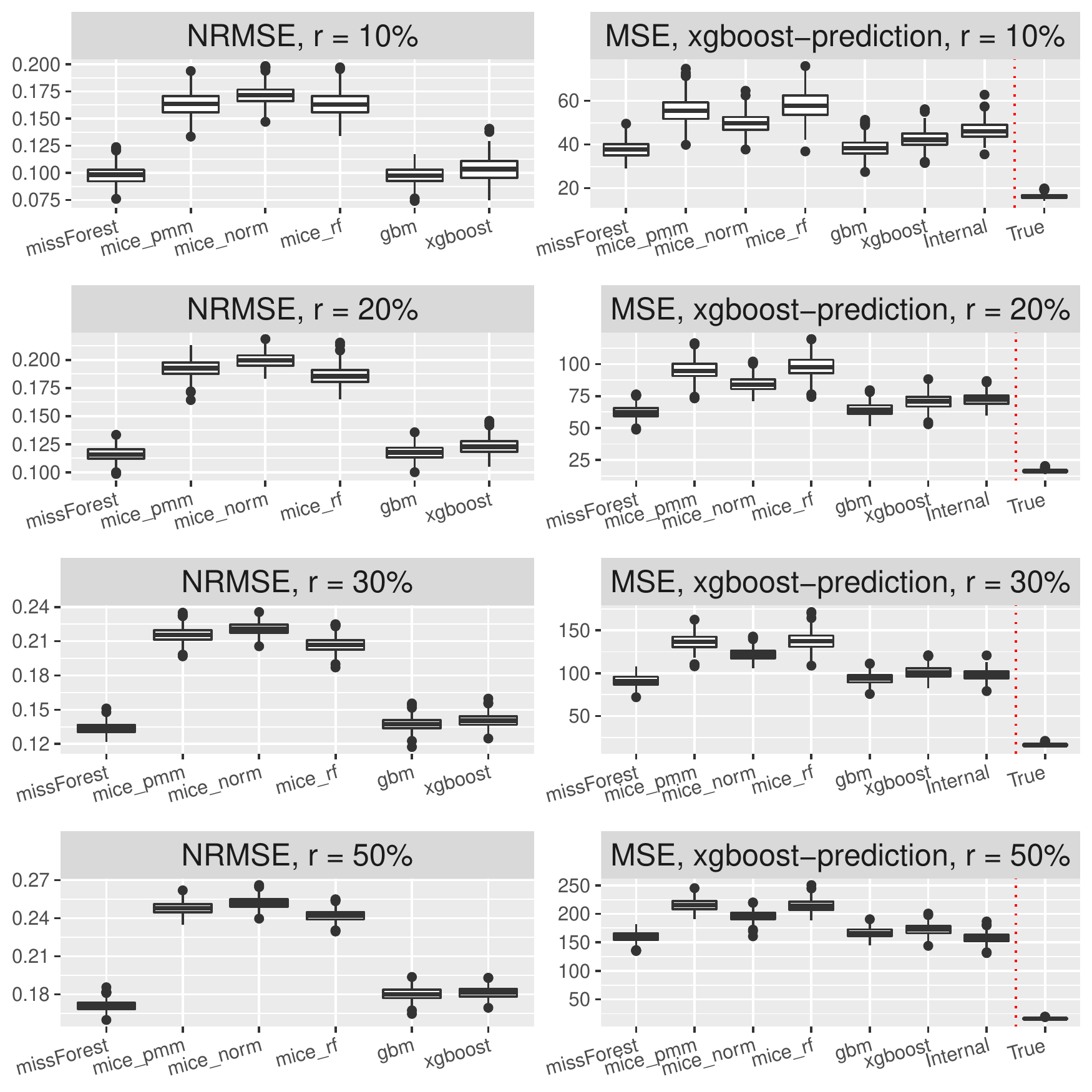}
	\captionof{figure}{Imputation (left column) and prediction accuracy (right column) measured by $NRMSE$ and $MSE$ using the XGBoost method for predicting scaled sound pressure in the Concrete dataset under various missing rates (listed row-wise). The $MSE$ is estimated based on a five-fold cross-validation procedure on the imputed dataset. The boxplots are over $500$ Monte-Carlo iterates. \textit{True} refers to the Concrete dataset without any missing values and a fitted Random Forest on the complete dataset. \textit{Internal} is the internal missing value treatment of the XGBoost method without prior imputation.}\label{Concrete_Plot_xgboost}
\end{center}

\subsection{QSAR dataset}

\begin{center}
	\centering
	\includegraphics[width=5in]{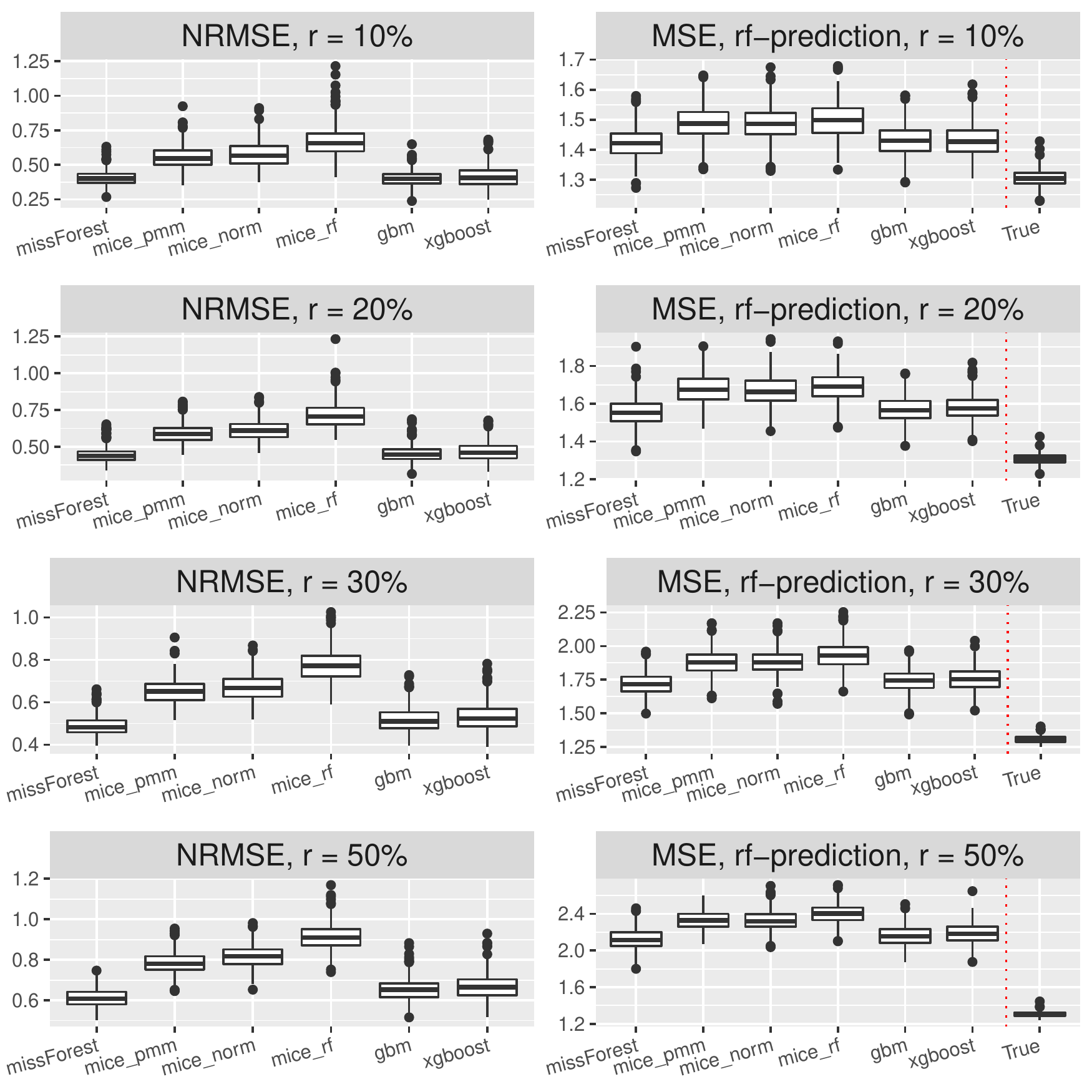}
	\captionof{figure}{Imputation (left column) and prediction accuracy (right column) measured by $NRMSE$ and $MSE$ using the Random Forest method for predicting scaled sound pressure in the QSAR dataset under various missing rates (listed row-wise). The $MSE$ is estimated based on a five-fold cross-validation procedure on the imputed dataset. The boxplots are over $500$ Monte-Carlo iterates. \textit{True} refers to the Airfoile dataset without any missing values and a fitted Random Forest on the complete dataset. }\label{QSAR_Plot_RF}
\end{center}

\begin{center}
	\centering
	\includegraphics[width=5in]{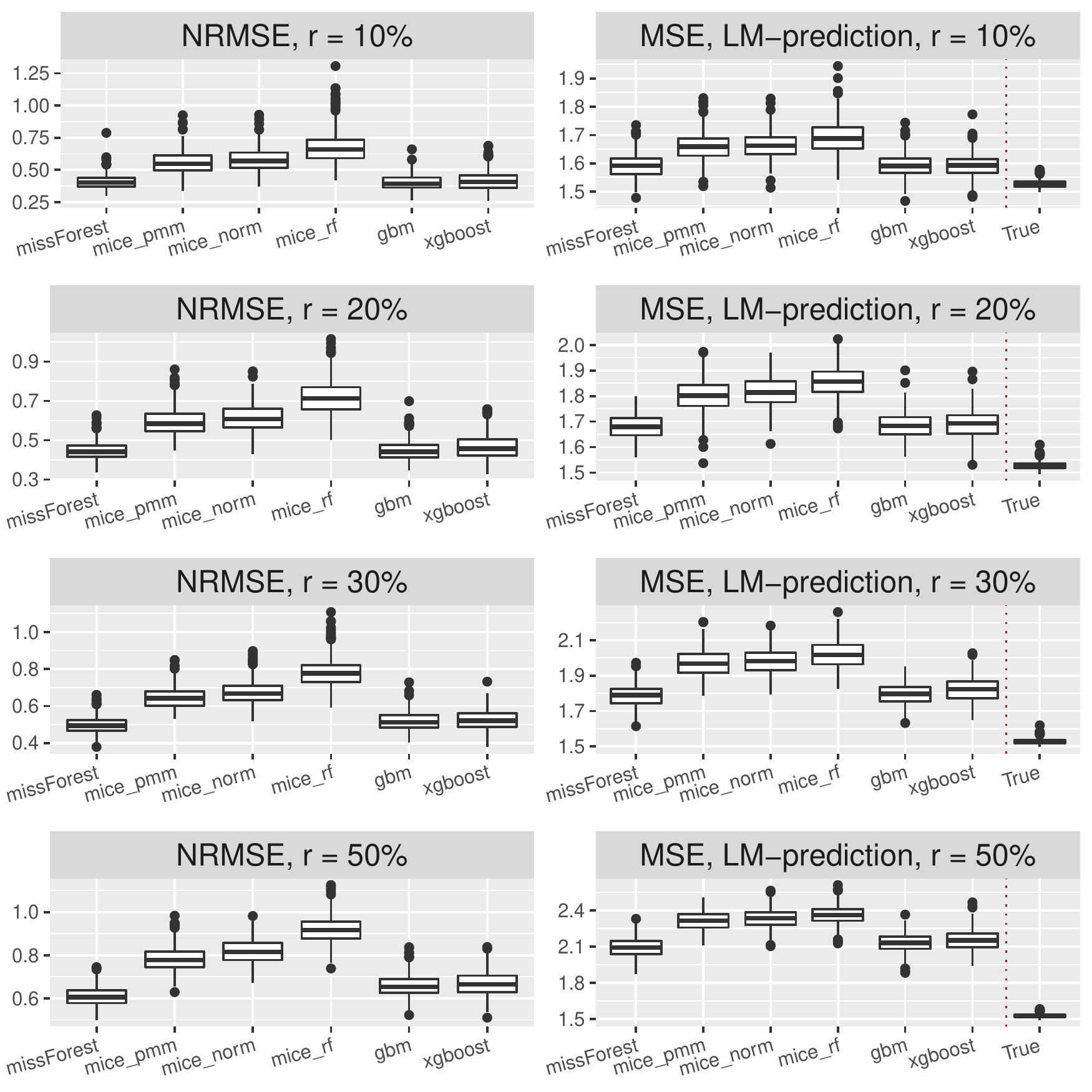}
	\captionof{figure}{Imputation (left column) and prediction accuracy (right column) measured by $NRMSE$ and $MSE$ using a linear model for predicting scaled sound pressure in the QSAR dataset under various missing rates (listed row-wise). The $MSE$ is estimated based on a five-fold cross-validation procedure on the imputed dataset. The boxplots are over $500$ Monte-Carlo iterates. \textit{True} refers to the QSAR dataset without any missing values and a fitted Random Forest on the complete dataset.}\label{Concrete_Plot_linear}
\end{center}

\begin{center}
	\centering
	\includegraphics[width=5in]{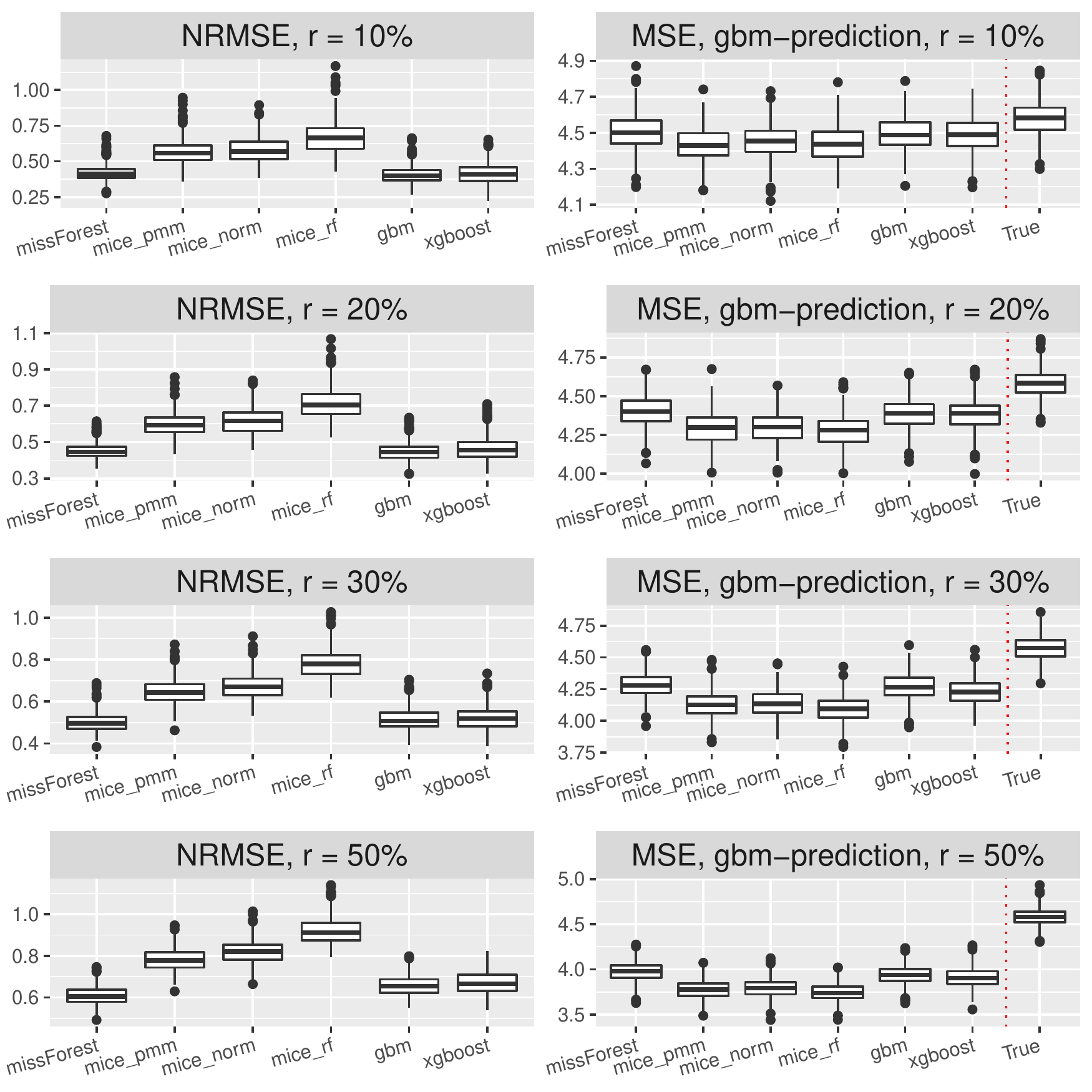}
	\captionof{figure}{Imputation (left column) and prediction accuracy (right column) measured by $NRMSE$ and $MSE$ using the SGB method for predicting scaled sound pressure in the QSAR dataset under various missing rates (listed row-wise). The $MSE$ is estimated based on a five-fold cross-validation procedure on the imputed dataset. The boxplots are over $500$ Monte-Carlo iterates. \textit{True} refers to the QSAR dataset without any missing values and a fitted Random Forest on the complete dataset.}\label{QSAR_Plot_gbm}
\end{center}

\begin{center}
	\centering
	\includegraphics[width=5in]{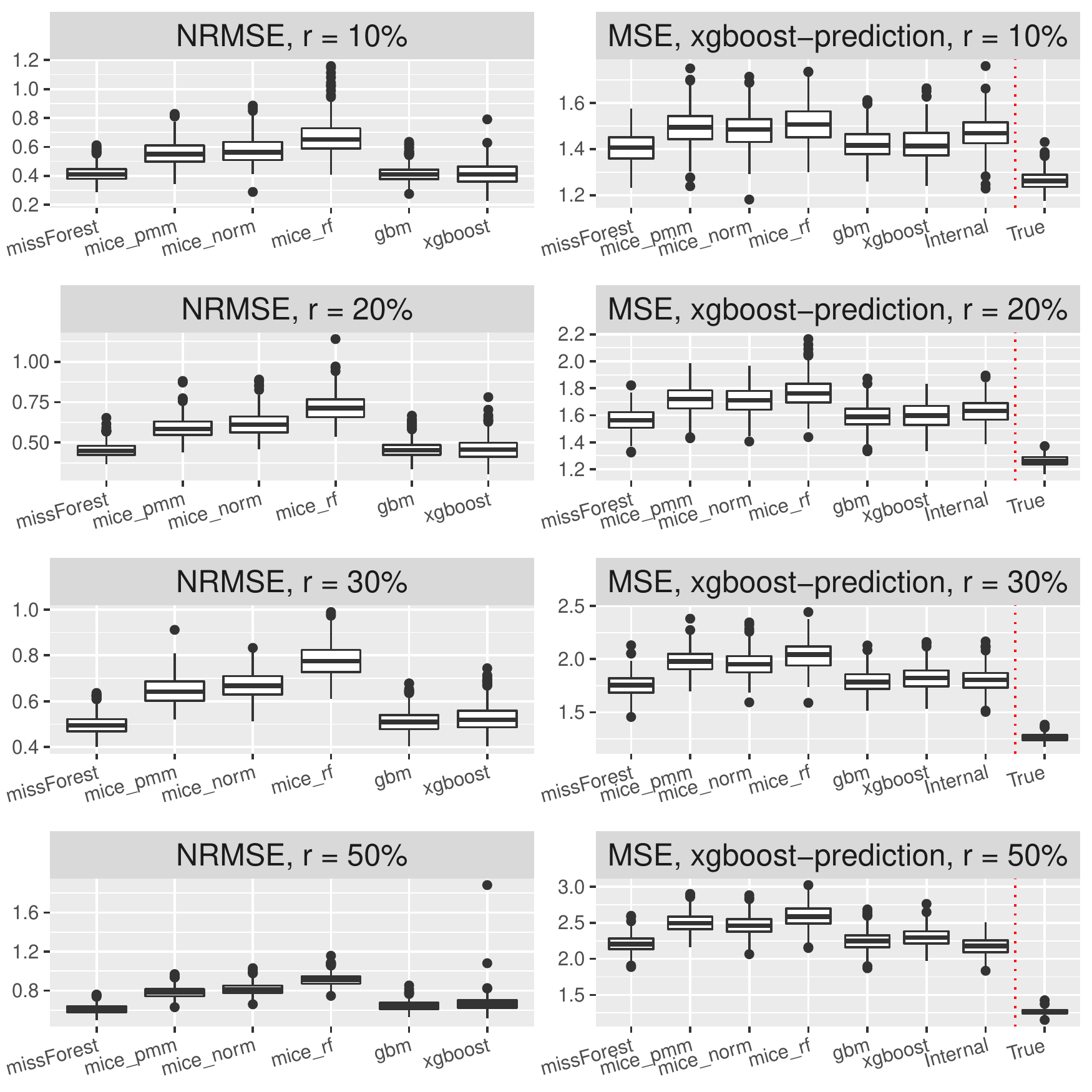}
	\captionof{figure}{Imputation (left column) and prediction accuracy (right column) measured by $NRMSE$ and $MSE$ using the XGBoost method for predicting scaled sound pressure in the QSAR dataset under various missing rates (listed row-wise). The $MSE$ is estimated based on a five-fold cross-validation procedure on the imputed dataset. The boxplots are over $500$ Monte-Carlo iterates. \textit{True} refers to the QSAR dataset without any missing values and a fitted Random Forest on the complete dataset. \textit{Internal} is the internal missing value treatment of the XGBoost method without prior imputation.}\label{QSAR_Plot_xgboost}
\end{center}

\subsection{Real Estate dataset}

\begin{center}
	\centering
	\includegraphics[width=5in]{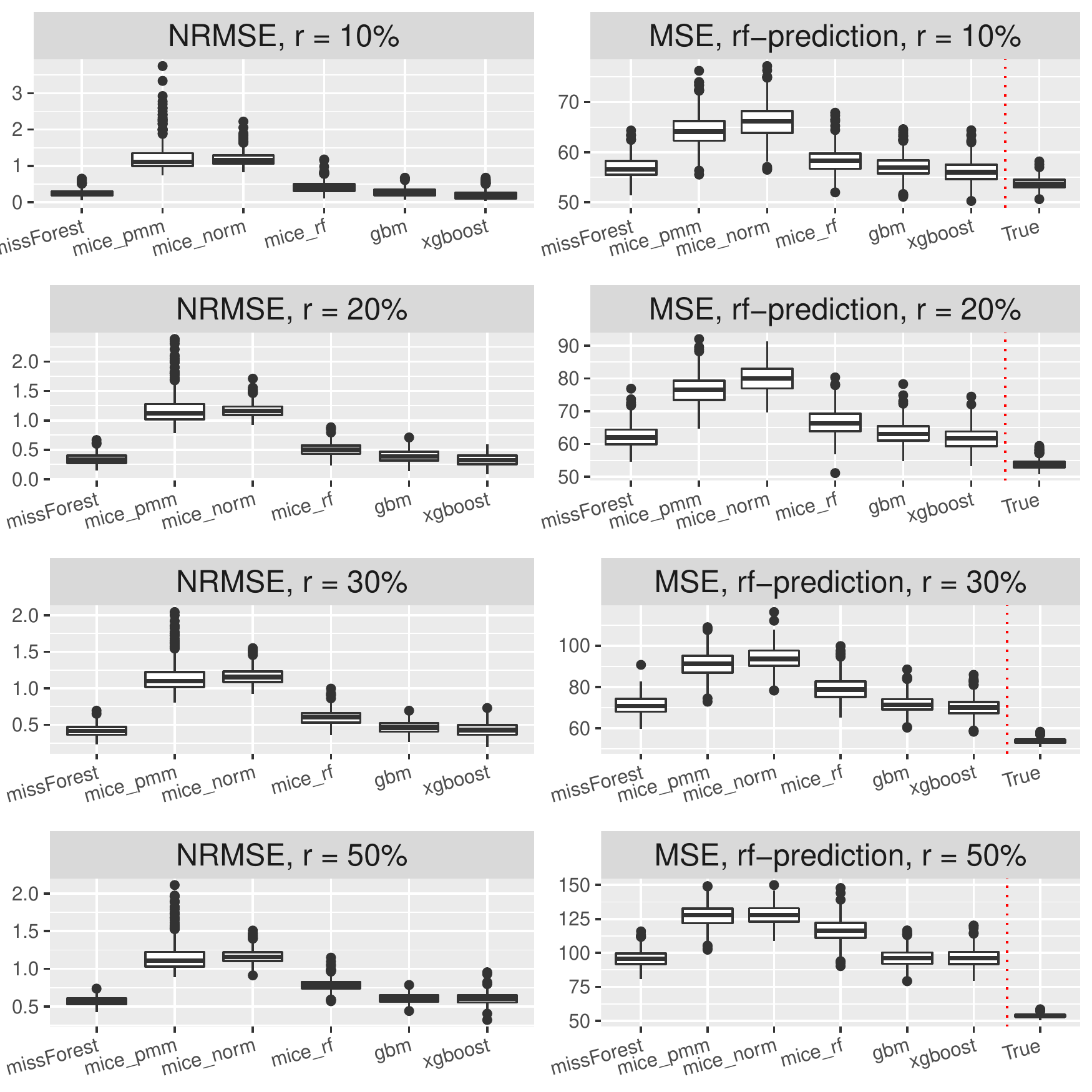}
	\captionof{figure}{Imputation (left column) and prediction accuracy (right column) measured by $NRMSE$ and $MSE$ using the Random Forest method for predicting scaled sound pressure in the Real Estate dataset under various missing rates (listed row-wise). The $MSE$ is estimated based on a five-fold cross-validation procedure on the imputed dataset. The boxplots are over $500$ Monte-Carlo iterates. \textit{True} refers to the Real Estate dataset without any missing values and a fitted Random Forest on the complete dataset. }\label{RealEstate_Plot_RF}
\end{center}

\begin{center}
	\centering
	\includegraphics[width=5in]{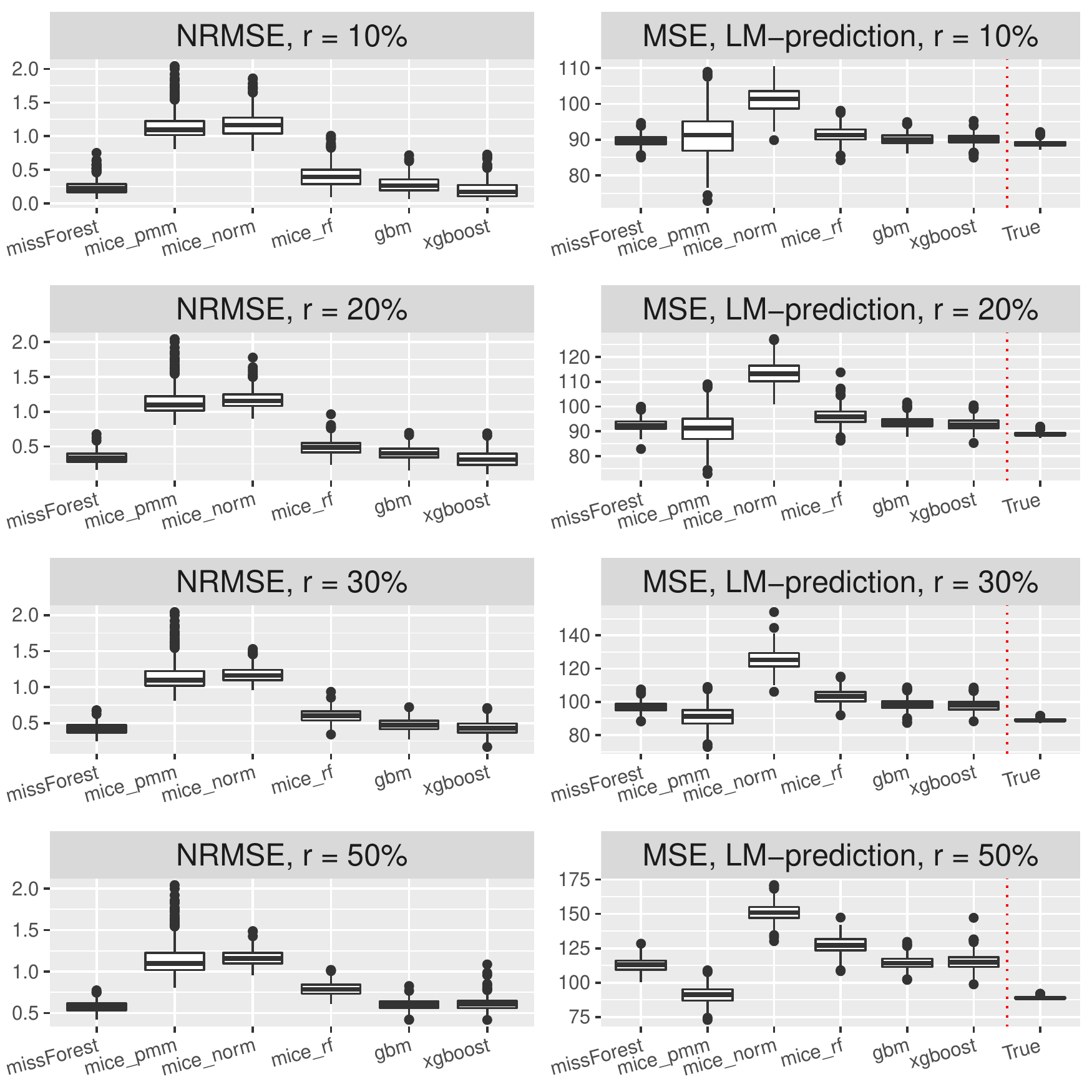}
	\captionof{figure}{Imputation (left column) and prediction accuracy (right column) measured by $NRMSE$ and $MSE$ using a linear model for predicting scaled sound pressure in the Real Estate dataset under various missing rates (listed row-wise). The $MSE$ is estimated based on a five-fold cross-validation procedure on the imputed dataset. The boxplots are over $500$ Monte-Carlo iterates. \textit{True} refers to the Real Estate dataset without any missing values and a fitted Random Forest on the complete dataset.}\label{RealEstate_Plot_linear}
\end{center}

\begin{center}
	\centering
	\includegraphics[width=5in]{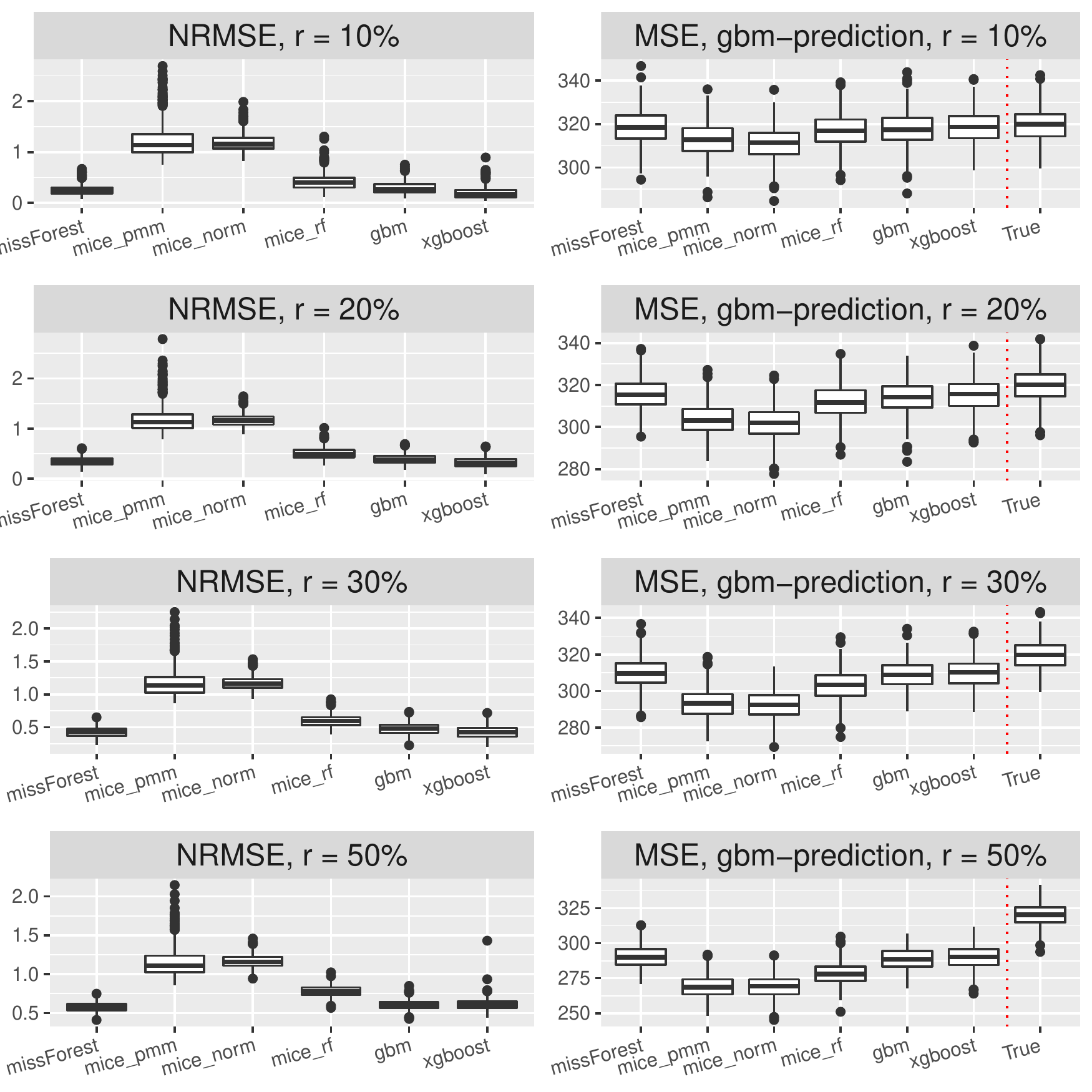}
	\captionof{figure}{Imputation (left column) and prediction accuracy (right column) measured by $NRMSE$ and $MSE$ using the SGB method for predicting scaled sound pressure in the Real Estate dataset under various missing rates (listed row-wise). The $MSE$ is estimated based on a five-fold cross-validation procedure on the imputed dataset. The boxplots are over $500$ Monte-Carlo iterates. \textit{True} refers to the Real Estate dataset without any missing values and a fitted Random Forest on the complete dataset.}\label{RealEstate_Plot_gbm}
\end{center}

\begin{center}
	\centering
	\includegraphics[width=5in]{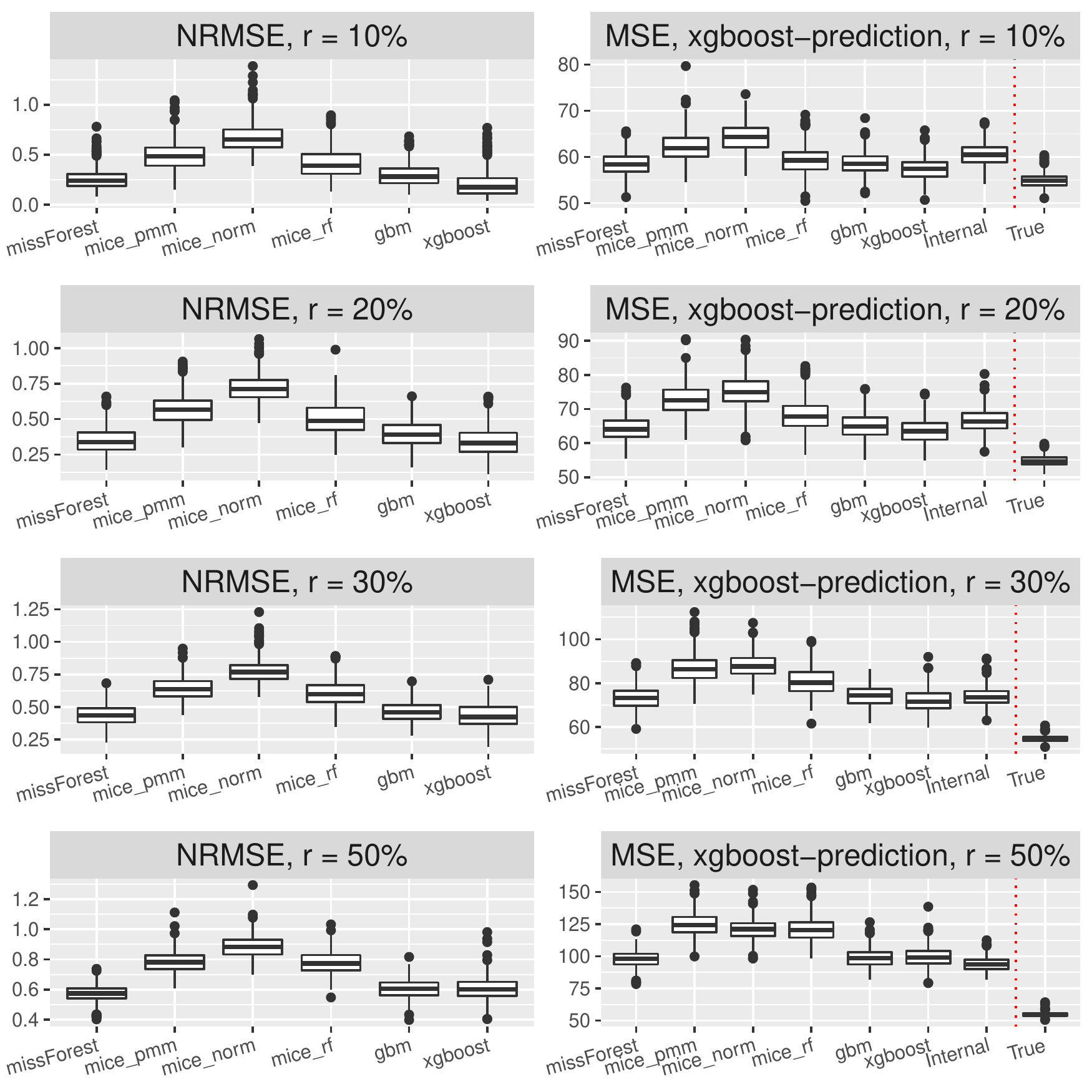}
	\captionof{figure}{Imputation (left column) and prediction accuracy (right column) measured by $NRMSE$ and $MSE$ using the XGBoost method for predicting scaled sound pressure in the Real Estate dataset under various missing rates (listed row-wise). The $MSE$ is estimated based on a five-fold cross-validation procedure on the imputed dataset. The boxplots are over $500$ Monte-Carlo iterates. \textit{True} refers to the Real Estate dataset without any missing values and a fitted Random Forest on the complete dataset. \textit{Internal} is the internal missing value treatment of the XGBoost method without prior imputation.}\label{RealEstate_Plot_xgboost}
\end{center}

\subsection{Power Plant dataset}

\begin{center}
	\centering
	\includegraphics[width=5in]{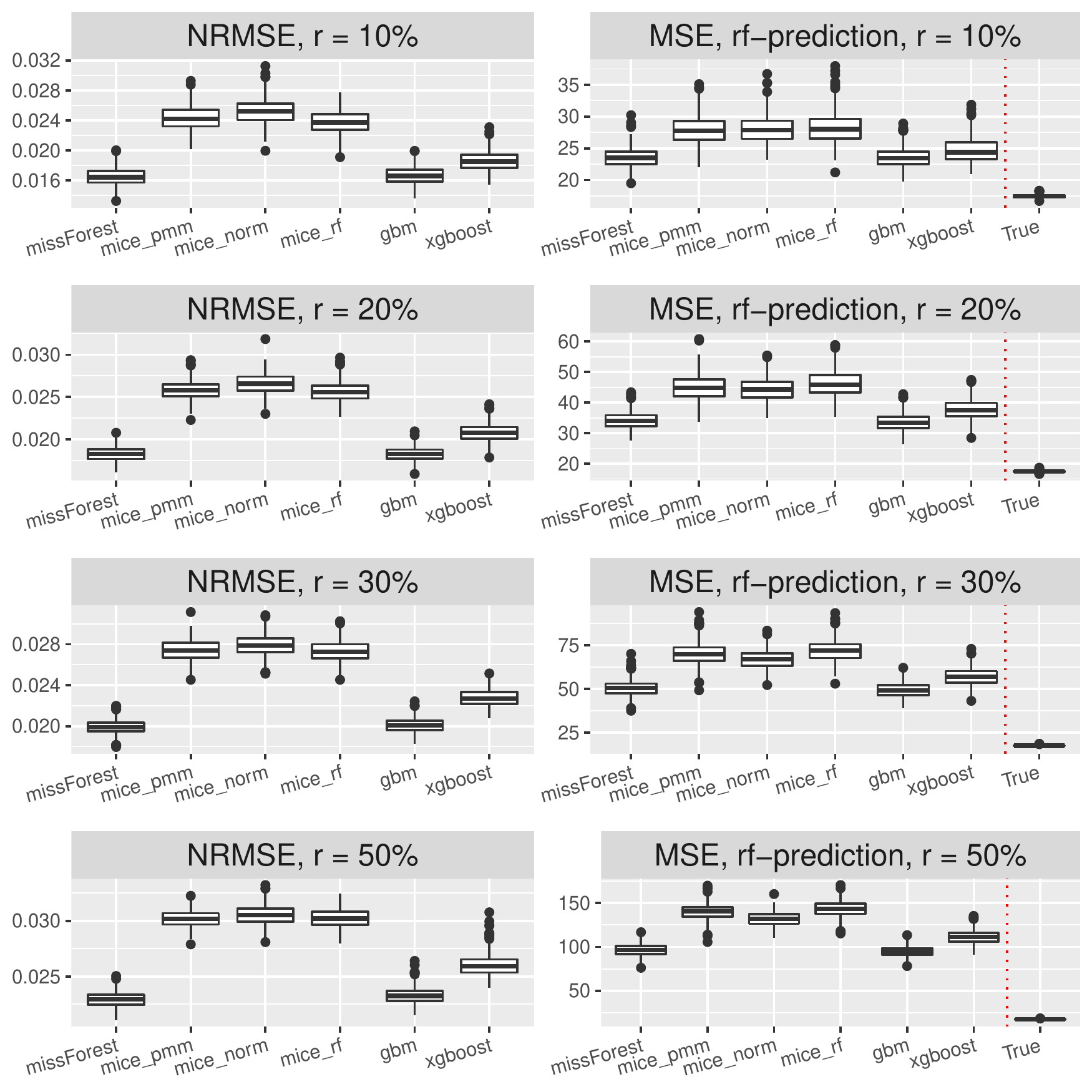}
	\captionof{figure}{Imputation (left column) and prediction accuracy (right column) measured by $NRMSE$ and $MSE$ using the Random Forest method for predicting scaled sound pressure in the Power Plant dataset under various missing rates (listed row-wise). The $MSE$ is estimated based on a five-fold cross-validation procedure on the imputed dataset. The boxplots are over $500$ Monte-Carlo iterates. \textit{True} refers to the Power Plant dataset without any missing values and a fitted Random Forest on the complete dataset. }\label{PowerPlant_Plot_RF}
\end{center}

\begin{center}
	\centering
	\includegraphics[width=5in]{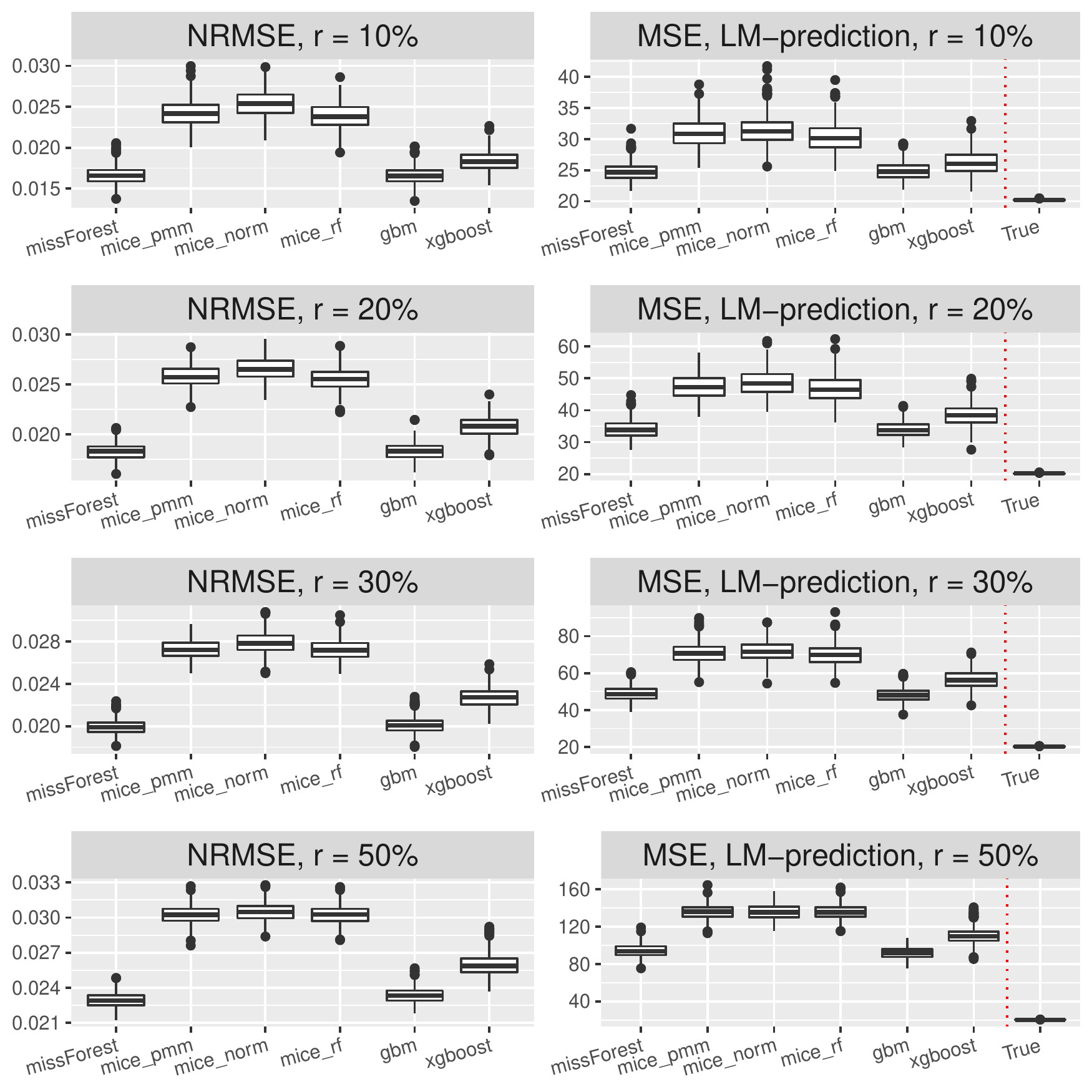}
	\captionof{figure}{Imputation (left column) and prediction accuracy (right column) measured by $NRMSE$ and $MSE$ using a linear model for predicting scaled sound pressure in the Power Plant dataset under various missing rates (listed row-wise). The $MSE$ is estimated based on a five-fold cross-validation procedure on the imputed dataset. The boxplots are over $500$ Monte-Carlo iterates. \textit{True} refers to the Power Plant dataset without any missing values and a fitted Random Forest on the complete dataset.}\label{PowerPlant_Plot_linear}
\end{center}

\begin{center}
	\centering
	\includegraphics[width=5in]{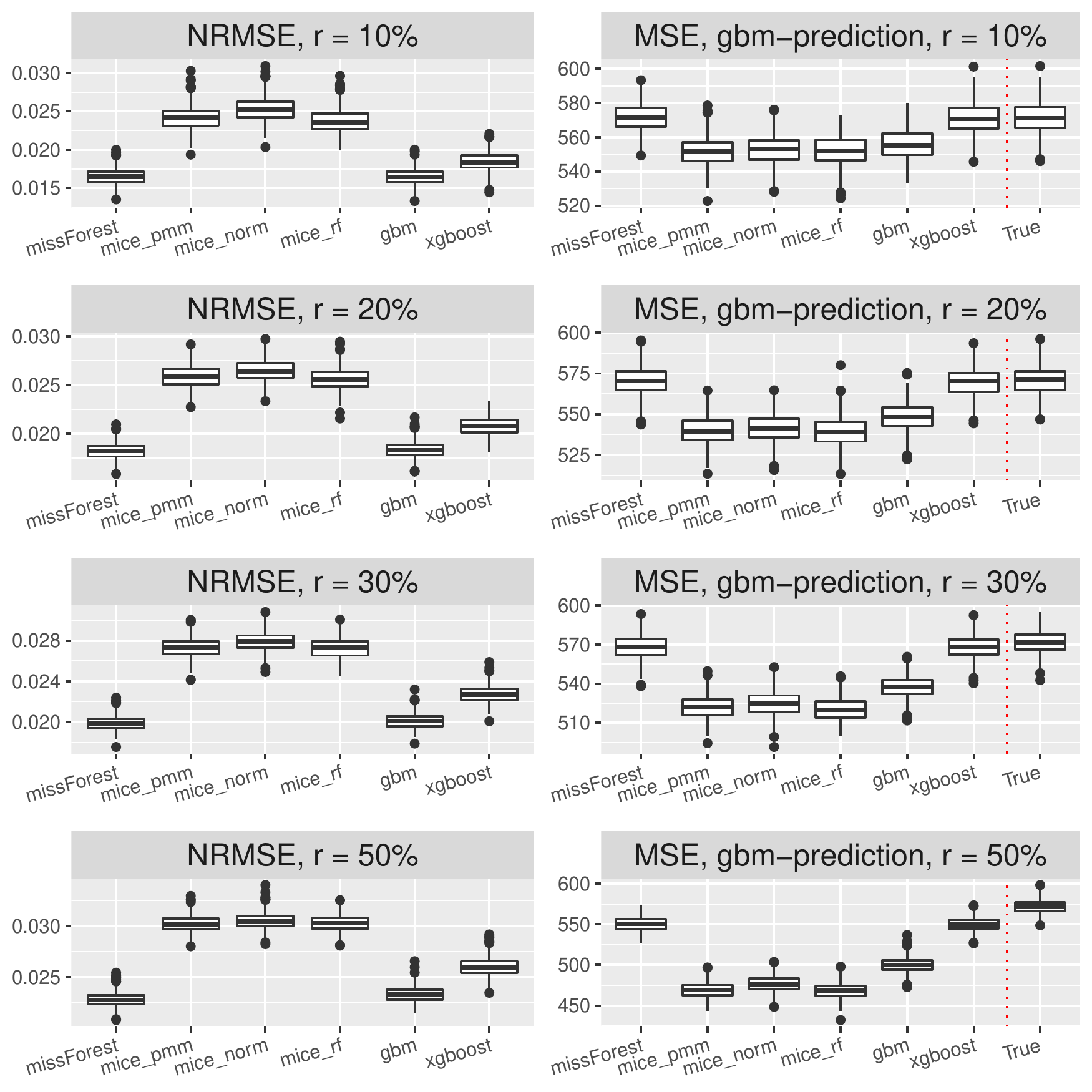}
	\captionof{figure}{Imputation (left column) and prediction accuracy (right column) measured by $NRMSE$ and $MSE$ using the SGB method for predicting scaled sound pressure in the Power Plant dataset under various missing rates (listed row-wise). The $MSE$ is estimated based on a five-fold cross-validation procedure on the imputed dataset. The boxplots are over $500$ Monte-Carlo iterates. \textit{True} refers to the Power Plant dataset without any missing values and a fitted Random Forest on the complete dataset.}\label{PowerPlant_Plot_gbm}
\end{center}

\begin{center}
	\centering
	\includegraphics[width=5in]{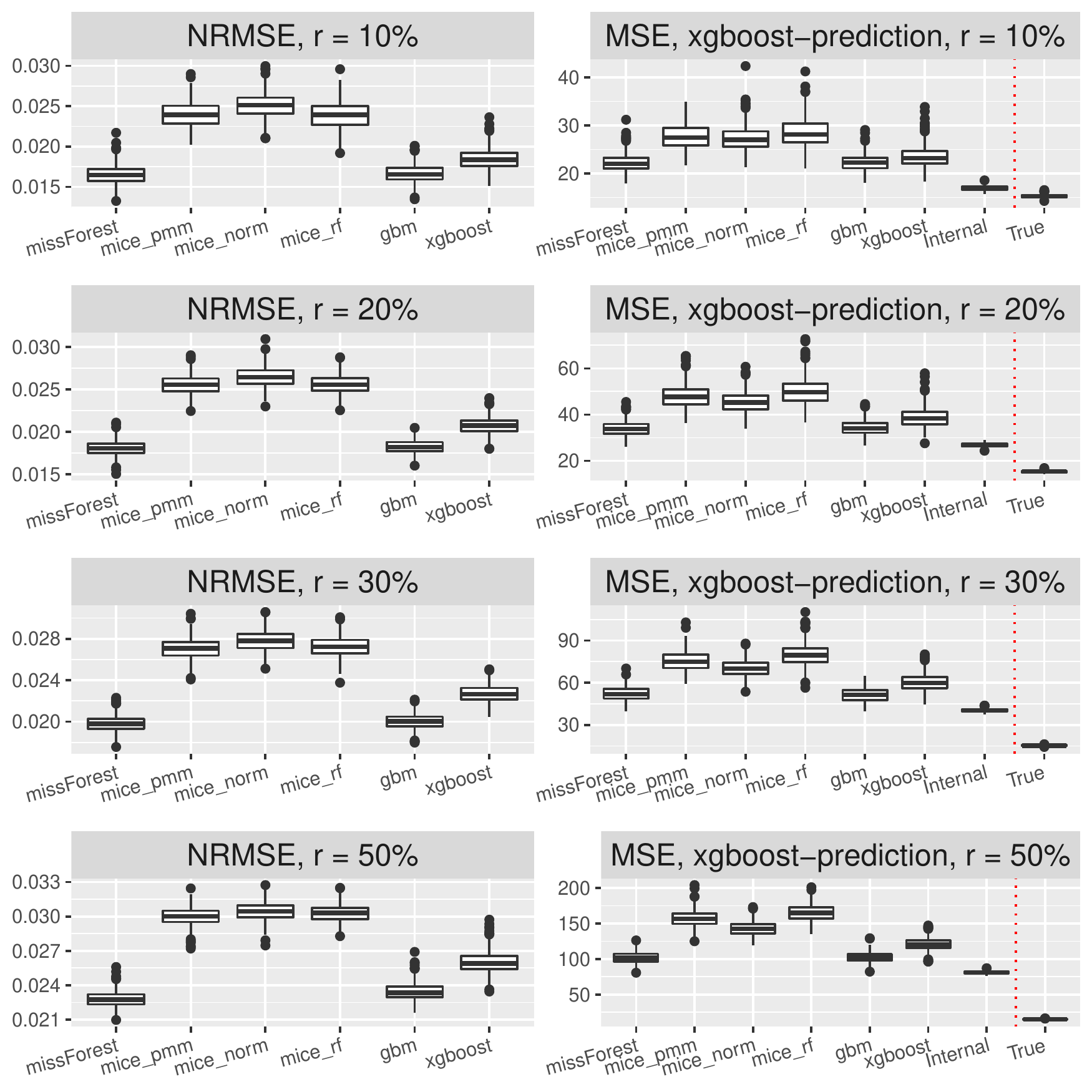}
	\captionof{figure}{Imputation (left column) and prediction accuracy (right column) measured by $NRMSE$ and $MSE$ using the XGBoost method for predicting scaled sound pressure in the Power Plant dataset under various missing rates (listed row-wise). The $MSE$ is estimated based on a five-fold cross-validation procedure on the imputed dataset. The boxplots are over $500$ Monte-Carlo iterates. \textit{True} refers to the Power Plant dataset without any missing values and a fitted Random Forest on the complete dataset. \textit{Internal} is the internal missing value treatment of the XGBoost method without prior imputation. }\label{PowerPlant_Plot_xgboost}
\end{center}

\section{Results on Prediction Coverage and Length}

\subsection{Linear model}

\begin{center}
	\centering
	\includegraphics[width=5in]{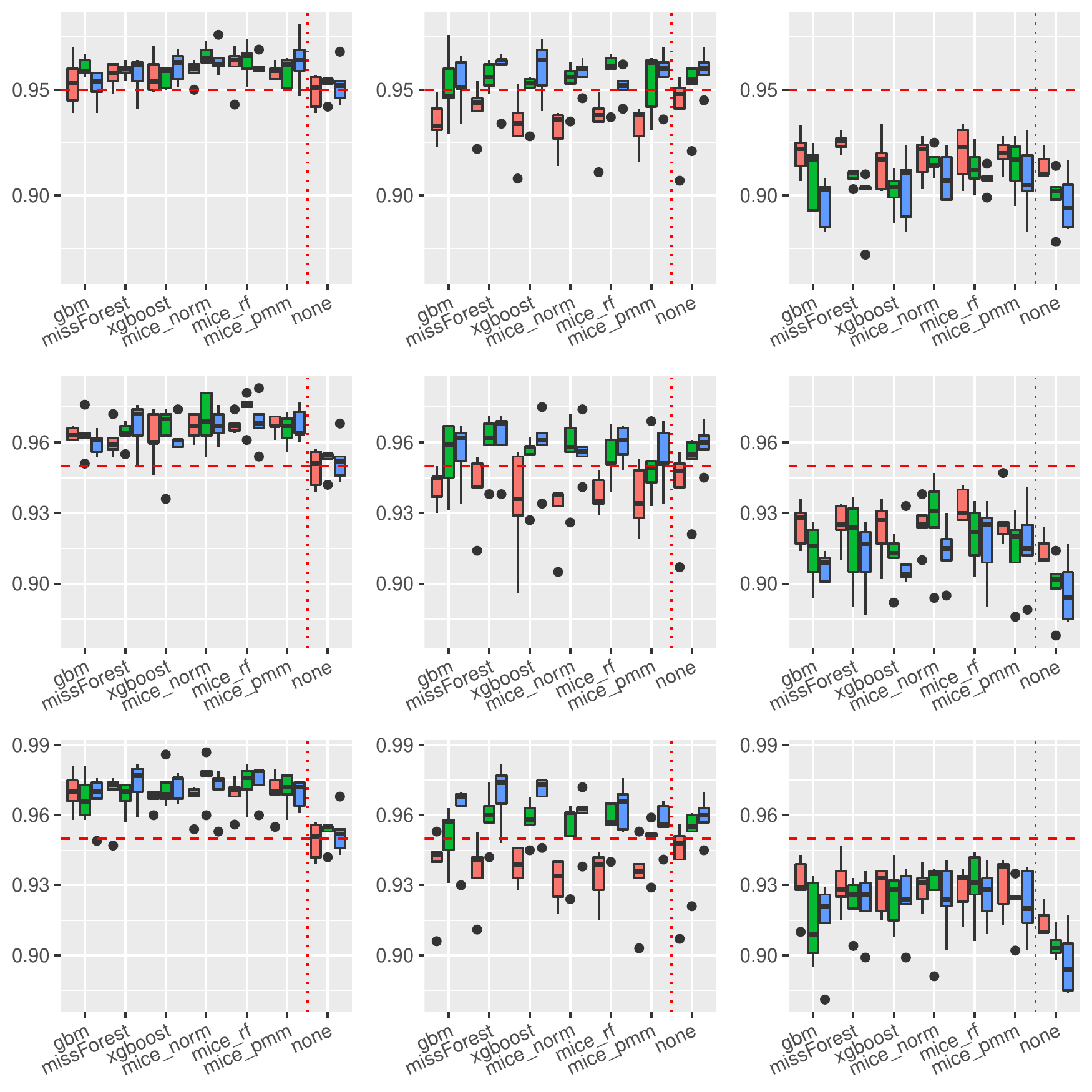}
	\captionof{figure}{ Boxplots of prediction  \textbf{coverage rates} under the \textbf{linear model}. The variation is over the different covariance structures of the features. Each row corresponds to one of the missing rates $r \in \{0.1, 0.2, 0.3\}$, while each column to the following prediction intervals: $PI_{QRF, n}$, $PI_{n, MCorrect}$ and the prediction interval based on the linear model. The tripple (red, green and blue) correspond to the sample sizes $n \in (100,500,1000)$.}\label{PI_Coverage_data_counter_linear}
\end{center}

\begin{center}
	\centering
	\includegraphics[width=5in]{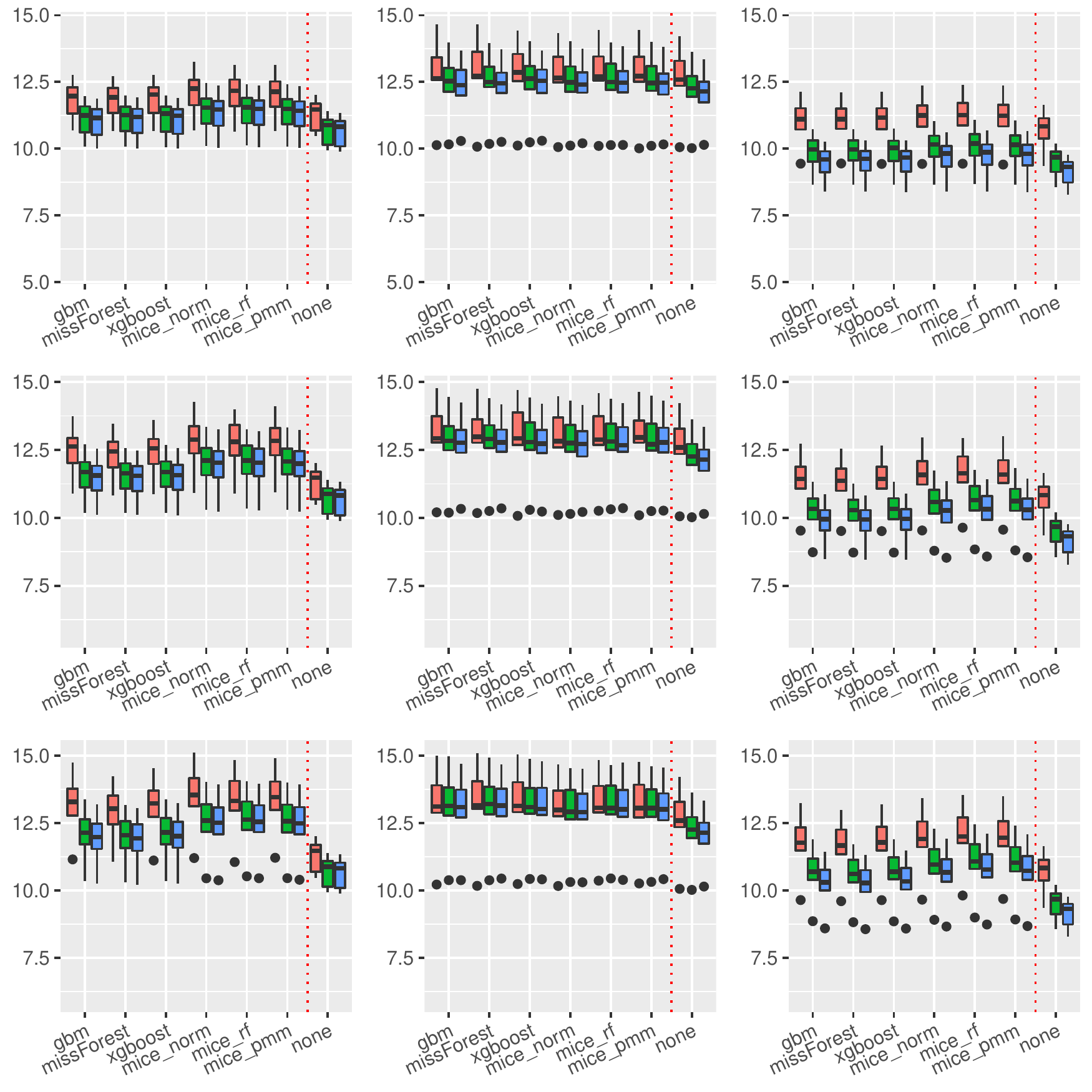}
	\captionof{figure}{Boxplots of prediction \textbf{interval lengths} under the \textbf{linear model}. The variation is over the different covariance structures of the features. Each row corresponds to one of the missing rates $r \in \{0.1, 0.2, 0.3\}$, while each column to the following prediction intervals: $PI_{QRF, n}$, $PI_{n, MCorrect}$ and the prediction interval based on the linear model. The tripple (red, green and blue) correspond to the sample sizes $n \in (100,500,1000)$.}\label{PI_Length_data_counter_linear}
\end{center}

\subsection{Trigonometric model}

\begin{center}
	\centering
	\includegraphics[width=5in]{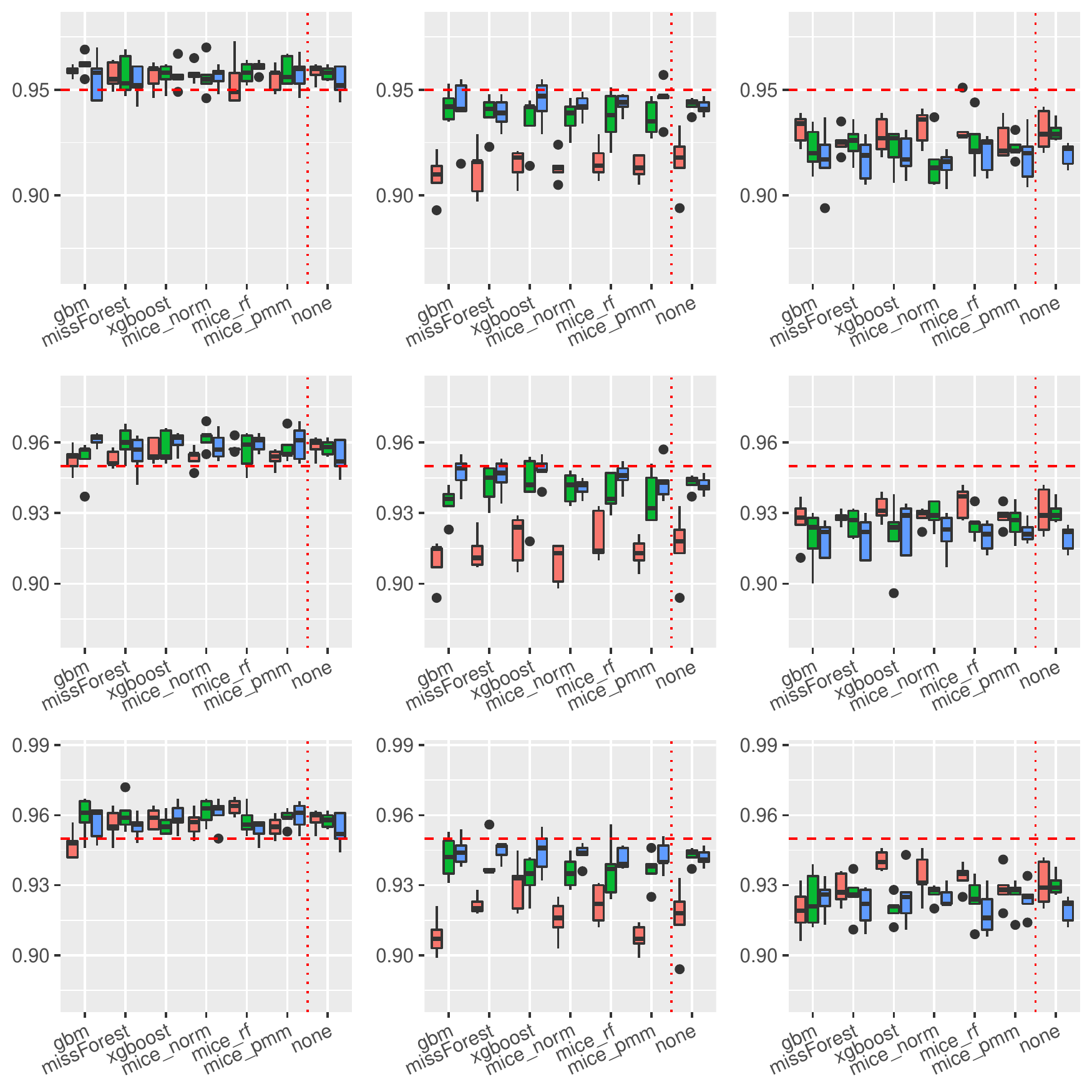}
	\captionof{figure}{ Boxplots of prediction  \textbf{coverage rates} under the \textbf{trigonometric model}. The variation is over the different covariance structures of the features. Each row corresponds to one of the missing rates $r \in \{0.1, 0.2, 0.3\}$, while each column to the following prediction intervals: $PI_{QRF, n}$, $PI_{n, MCorrect}$ and the prediction interval based on the linear model. The tripple (red, green and blue) correspond to the sample sizes $n \in (100,500,1000)$.}\label{PI_Coverage_data_counter_sinus}
\end{center}

\begin{center}
	\centering
	\includegraphics[width=5in]{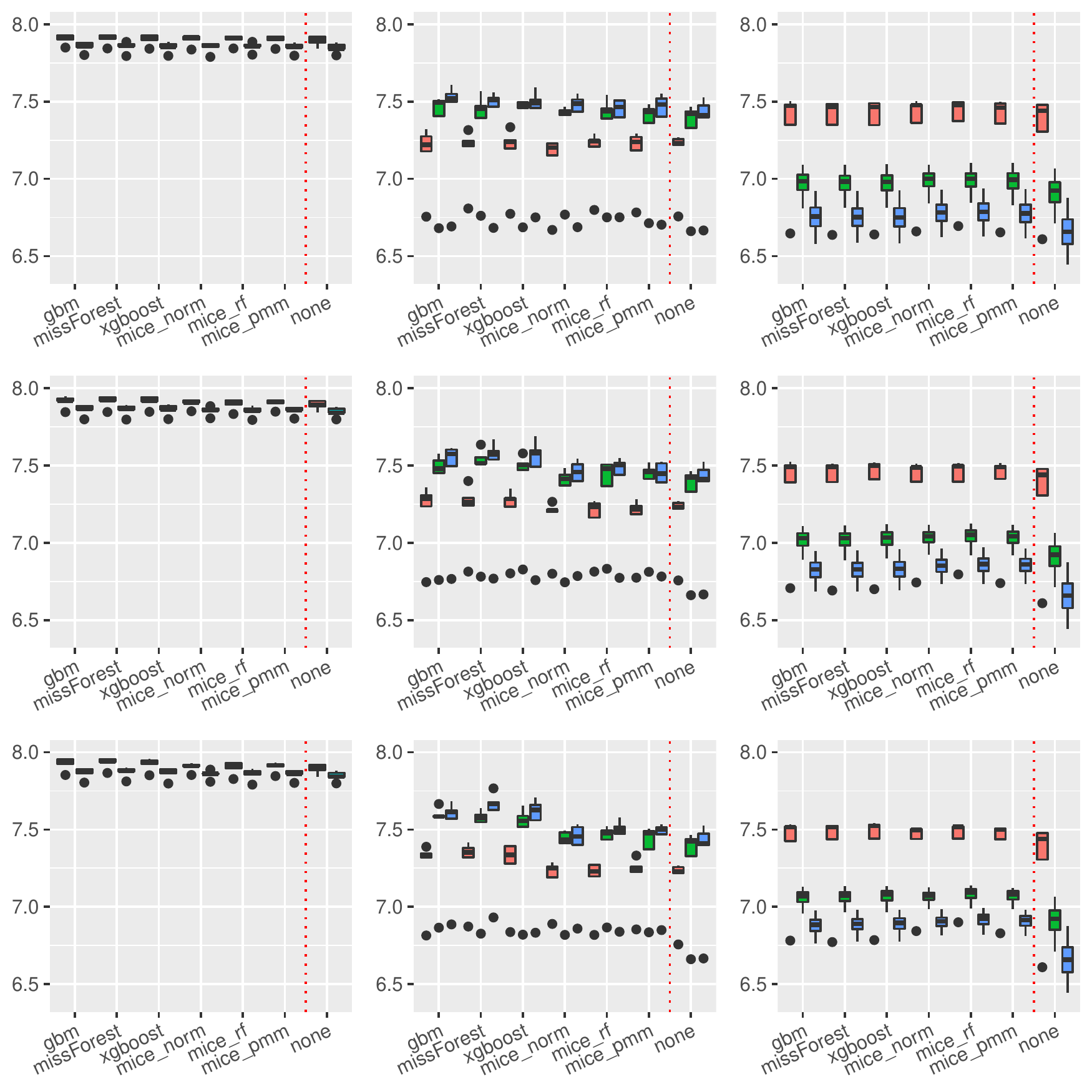}
	\captionof{figure}{Boxplots of prediction \textbf{interval lengths} under the \textbf{trigonometric model}. The variation is over the different covariance structures of the features. Each row corresponds to one of the missing rates $r \in \{0.1, 0.2, 0.3\}$, while each column to the following prediction intervals: $PI_{QRF, n}$, $PI_{n, MCorrect}$ and the prediction interval based on the linear model. The tripple (red, green and blue) correspond to the sample sizes $n \in (100,500,1000)$.}\label{PI_Length_data_counter_sinus}
\end{center}

\subsection{Polynomial model} 

\begin{center}
	\centering
	\includegraphics[width=5in]{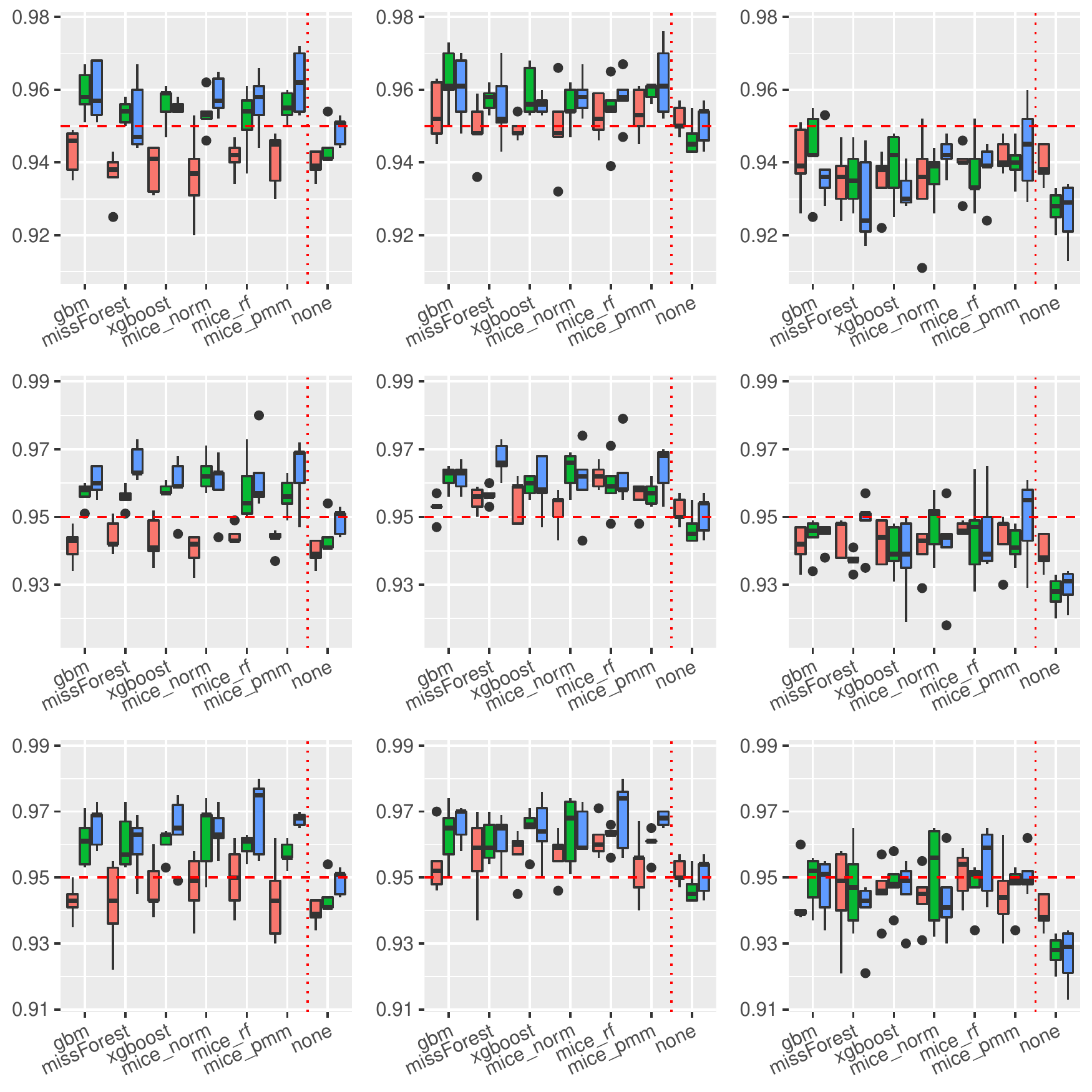}
	\captionof{figure}{ Boxplots of prediction  \textbf{coverage rates} under the \textbf{polynomial model}. The variation is over the different covariance structures of the features. Each row corresponds to one of the missing rates $r \in \{0.1, 0.2, 0.3\}$, while each column to the following prediction intervals: $PI_{n, empQ}$, $PI_{n, ResVar}$ and $PI_{n, weighted}$. The tripple (red, green and blue) correspond to the sample sizes $n \in (100,500,1000)$.}\label{PI_Coverage_data_optimal_polynom}
\end{center}

\begin{center}
	\centering
	\includegraphics[width=5in]{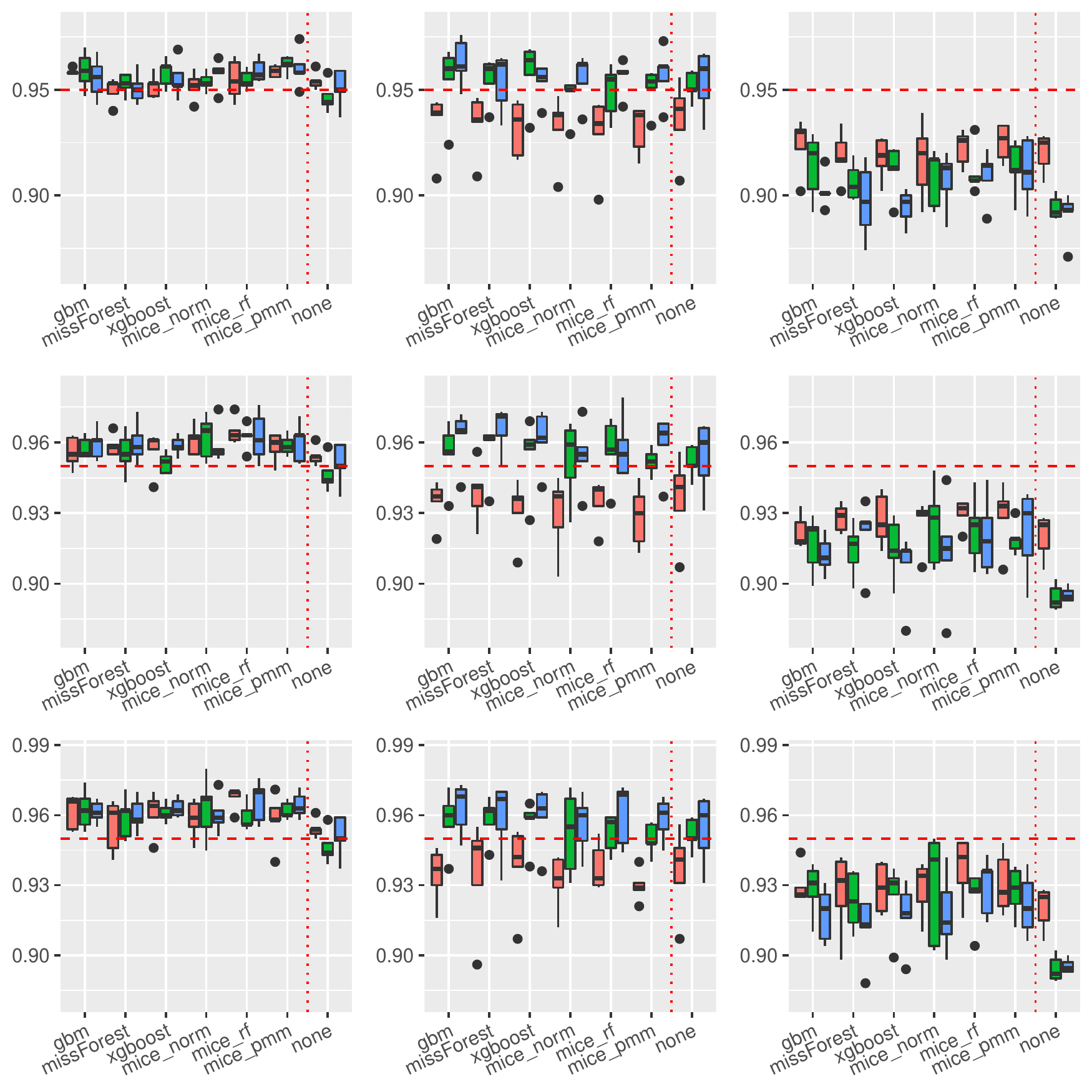}
	\captionof{figure}{ Boxplots of prediction  \textbf{coverage rates} under the \textbf{polynomial model}. The variation is over the different covariance structures of the features. Each row corresponds to one of the missing rates $r \in \{0.1, 0.2, 0.3\}$, while each column to the following prediction intervals: $PI_{QRF, n}$, $PI_{n, MCorrect}$ and the prediction interval based on the linear model. The tripple (red, green and blue) correspond to the sample sizes $n \in (100,500,1000)$.}\label{PI_Coverage_data_counter_polynom}
\end{center}

\begin{center}
	\centering
	\includegraphics[width=5in]{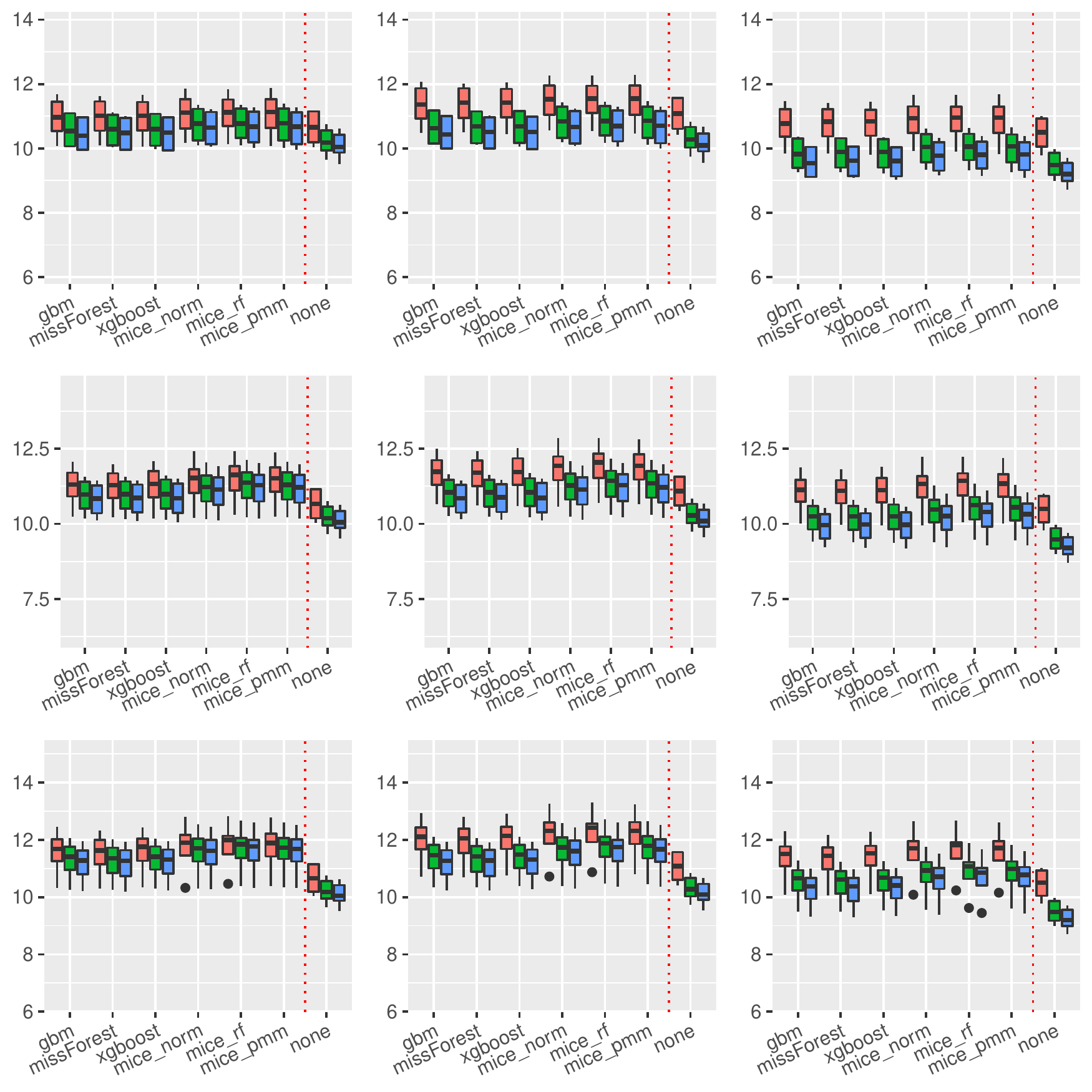}
	\captionof{figure}{Boxplots of prediction \textbf{interval lengths} under the \textbf{polynomial model}. The variation is over the different covariance structures of the features. Each row corresponds to one of the missing rates $r \in \{0.1, 0.2, 0.3\}$, while each column to the following prediction intervals: $PI_{n, empQ}$, $PI_{n, ResVar}$ and $PI_{n, weighted}$. The tripple (red, green and blue) correspond to the sample sizes $n \in (100,500,1000)$.}\label{PI_Length_data_optimal_polynom}
\end{center}

\begin{center}
	\centering
	\includegraphics[width=5in]{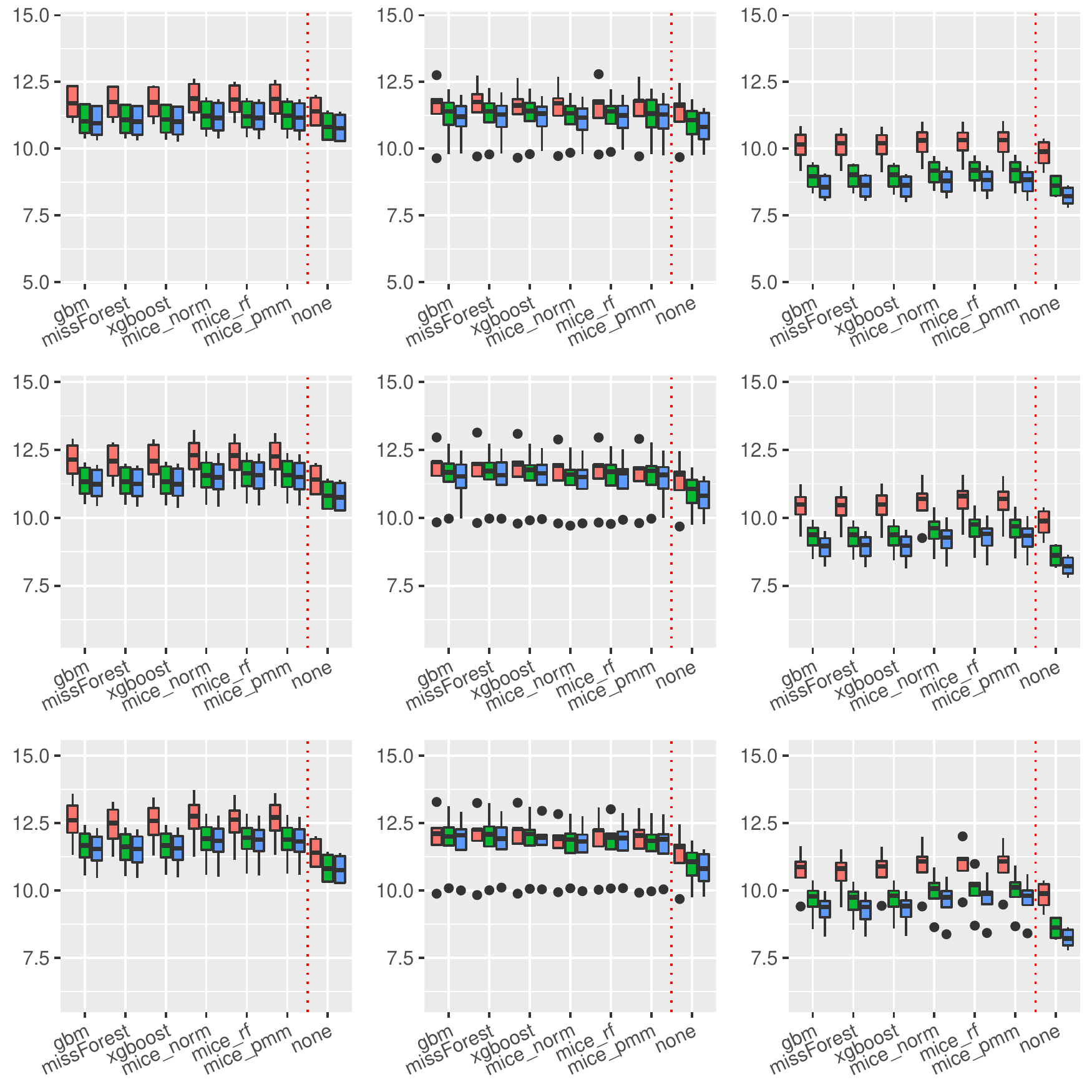}
	\captionof{figure}{Boxplots of prediction \textbf{interval lengths} under the \textbf{polynomial model}. The variation is over the different covariance structures of the features. Each row corresponds to one of the missing rates $r \in \{0.1, 0.2, 0.3\}$, while each column to the following prediction intervals: $PI_{QRF, n}$, $PI_{n, MCorrect}$ and the prediction interval based on the linear model. The tripple (red, green and blue) correspond to the sample sizes $n \in (100,500,1000)$.}\label{PI_Length_data_counter_polynom}
\end{center}

\subsection{Non-continuous model}

\begin{center}
	\centering
	\includegraphics[width=5in]{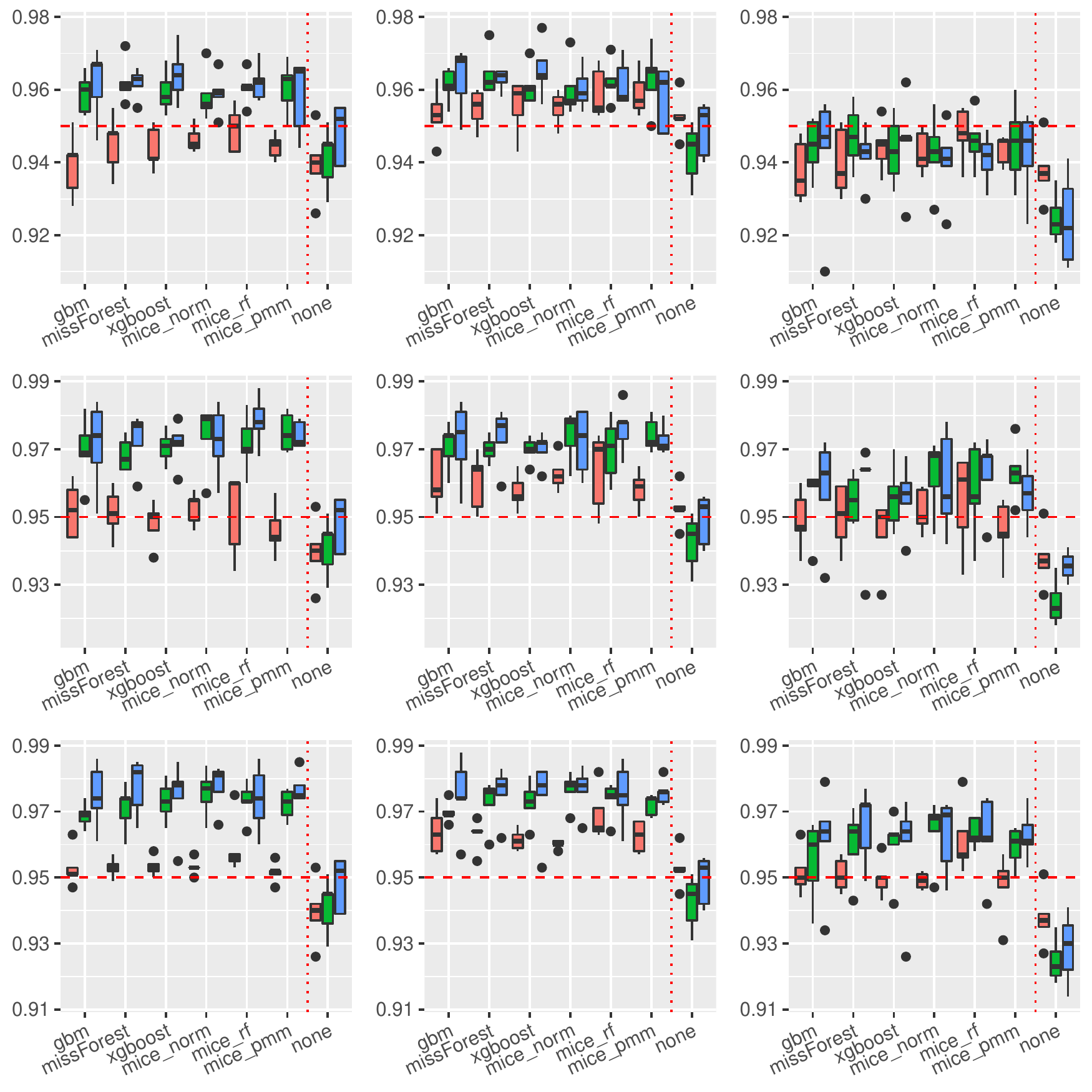}
	\captionof{figure}{ Boxplots of prediction  \textbf{coverage rates} under the \textbf{non-continuous model}. The variation is over the different covariance structures of the features. Each row corresponds to one of the missing rates $r \in \{0.1, 0.2, 0.3\}$, while each column to the following prediction intervals: $PI_{n, empQ}$, $PI_{n, ResVar}$ and $PI_{n, weighted}$. The tripple (red, green and blue) correspond to the sample sizes $n \in (100,500,1000)$.}\label{PI_Coverage_data_optimal_non-continuous}
\end{center}

\begin{center}
	\centering
	\includegraphics[width=5in]{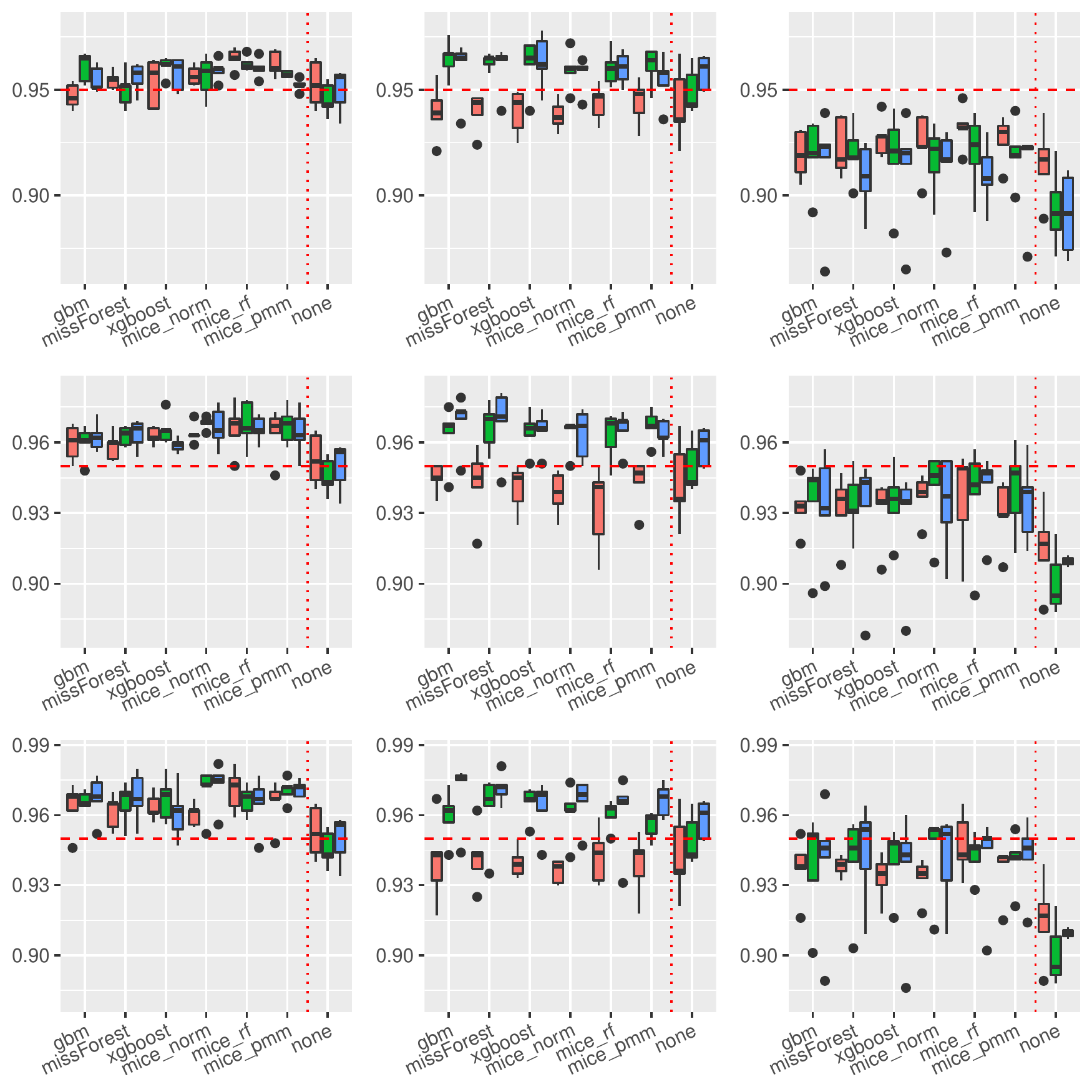}
	\captionof{figure}{ Boxplots of prediction  \textbf{coverage rates} under the \textbf{non-continuous model}. The variation is over the different covariance structures of the features. Each row corresponds to one of the missing rates $r \in \{0.1, 0.2, 0.3\}$, while each column to the following prediction intervals: $PI_{QRF, n}$, $PI_{n, MCorrect}$ and the prediction interval based on the linear model. The tripple (red, green and blue) correspond to the sample sizes $n \in (100,500,1000)$.}\label{PI_Coverage_data_counter_non-continuous}
\end{center}

\begin{center}
	\centering
	\includegraphics[width=5in]{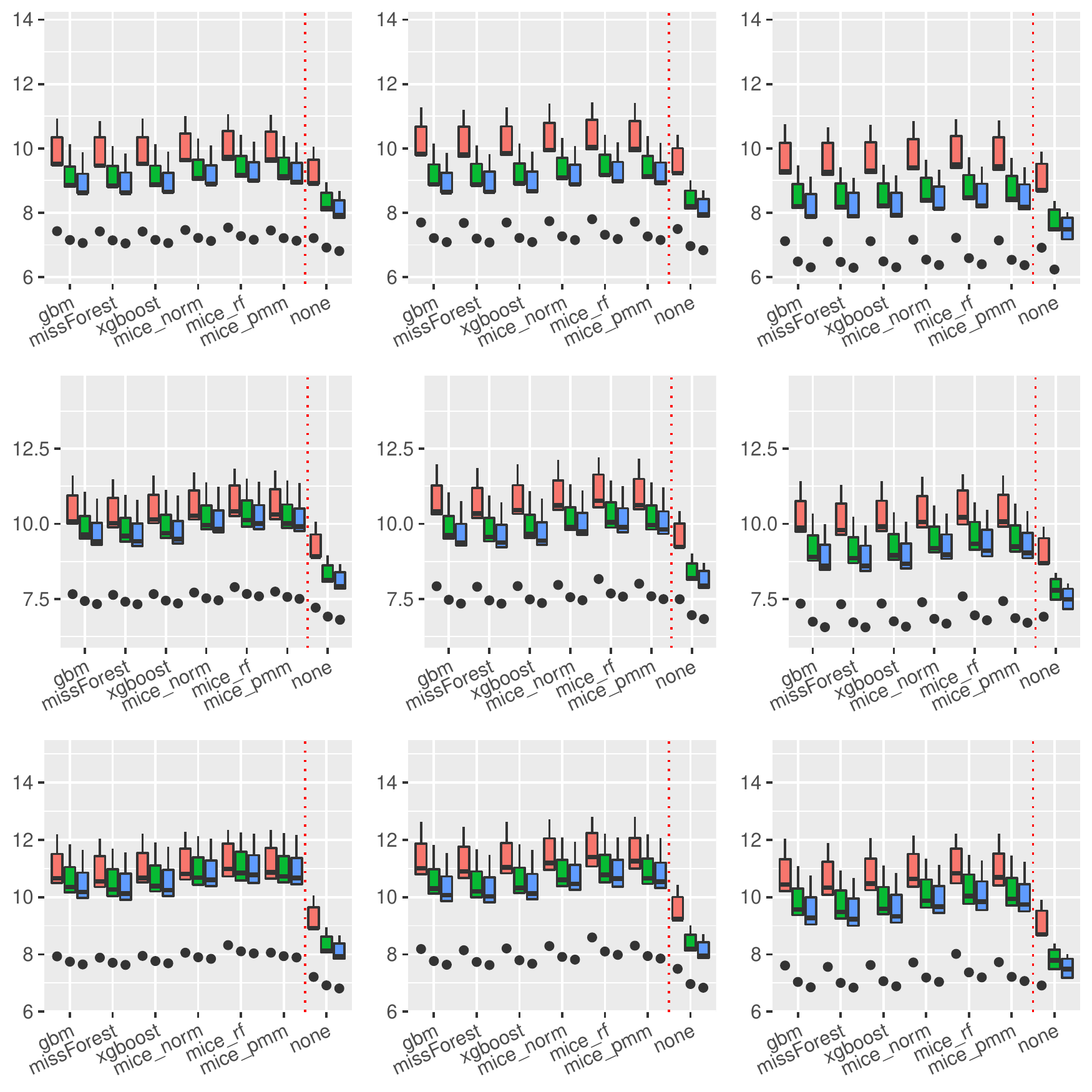}
	\captionof{figure}{Boxplots of prediction \textbf{interval lengths} under the \textbf{non-continuous model}. The variation is over the different covariance structures of the features. Each row corresponds to one of the missing rates $r \in \{0.1, 0.2, 0.3\}$, while each column to the following prediction intervals: $PI_{n, empQ}$, $PI_{n, ResVar}$ and $PI_{n, weighted}$. The tripple (red, green and blue) correspond to the sample sizes $n \in (100,500,1000)$.}\label{PI_Length_data_optimal_non-continuous}
\end{center}

\begin{center}
	\centering
	\includegraphics[width=5in]{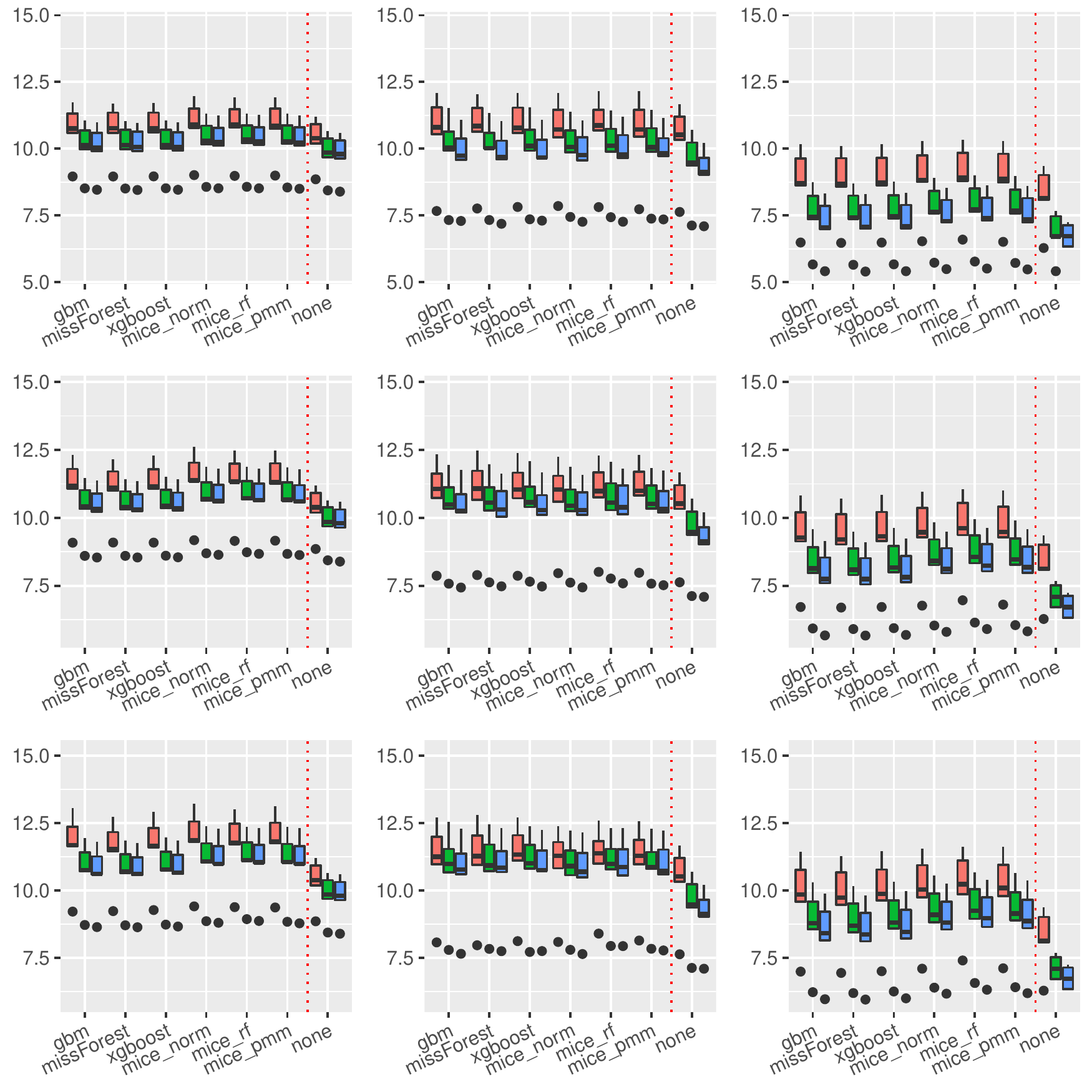}
	\captionof{figure}{Boxplots of prediction \textbf{interval lengths} under the \textbf{non-continuous model}. The variation is over the different covariance structures of the features. Each row corresponds to one of the missing rates $r \in \{0.1, 0.2, 0.3\}$, while each column to the following prediction intervals: $PI_{QRF, n}$, $PI_{n, MCorrect}$ and the prediction interval based on the linear model. The tripple (red, green and blue) correspond to the sample sizes $n \in (100,500,1000)$.}\label{PI_Length_data_counter_non-continuous}
\end{center}